\documentclass[final,5p,times,twocolumn]{elsarticle}

\usepackage{times}
\usepackage{epsfig}
\usepackage{graphicx}
\usepackage{amsmath}
\usepackage{amssymb}
\usepackage{epsfig}
\usepackage{commath}
\usepackage{tabu}
\usepackage{flexisym}
\usepackage{multirow}
\usepackage[table,xcdraw]{xcolor}
\colorlet{myblue}{black!45!black}
\colorlet{mymagenta2}{magenta!45!magenta}
\colorlet{myorange2}{orange!45!orange}
\colorlet{mygreen}{black!45!black}
\colorlet{mymagenta}{black!45!black}
\colorlet{myorange}{black!45!black}
\usepackage{caption}
\usepackage{subcaption}
\usepackage{comment}
\usepackage{fancyhdr}
\usepackage{scrextend}
\usepackage{setspace}
\usepackage{algorithm}
\usepackage{algpseudocode}
\usepackage[pagebackref=true,breaklinks=true,colorlinks,bookmarks=false]{hyperref}
\usepackage{algorithm} 
\usepackage{algpseudocode} 

\usepackage[utf8]{inputenc}
\usepackage[english]{babel}
\usepackage{fancyhdr}

\journal{Neural Networks}
\makeatletter
\def\ps@pprintTitle{%
 \let\@oddhead\@empty
 \let\@evenhead\@empty
 \def\@oddfoot{}%
 \let\@evenfoot\@oddfoot}
\makeatother

\begin{document}

\newcommand{\etal}{\textit{et al}. }
\newcommand{\ie}{\textit{i}.\textit{e}., }
\begin{frontmatter}
\pagestyle{fancy}
\title{HRel: Filter Pruning based on High Relevance between Activation Maps and Class Labels}
\author{CH Sarvani$^1$, Mrinmoy Ghorai$^1$, Shiv Ram Dubey$^2$, SH Shabbeer Basha$^3$}
\address{
$^1$Computer Vision Group, Indian Institute of Information Technology, Sri City, Chittoor, Andhra Pradesh- 517646, India.\\
$^2$Computer Vision and Biometrics Laboratory, Indian Institute of Information Technology, Allahabad, Uttar Pradesh- 211015, India. \\
$^3$DryvAmigo, Banglore, Karnataka, India.
\\[\bigskipamount]
{\texttt{\{sarvani.ch, mrinmoy.ghorai\}@iiits.in, srdubey@iiita.ac.in, shabbeer.sh@dryvamigo.com}}}

\begin{abstract}
This paper proposes an Information Bottleneck theory based filter pruning method that uses a statistical measure called Mutual Information (MI). The MI between filters and class labels, also called \textit{Relevance}, is computed using the filter's activation maps and the annotations. The filters having High Relevance (HRel) are considered to be more important. Consequently, the least important filters, which have lower Mutual Information with the class labels, are pruned. Unlike the existing MI based pruning methods, the proposed method determines the significance of the filters purely based on their corresponding activation map's relationship with the class labels. Architectures such as LeNet-5, VGG-16, ResNet-56\textcolor{myblue}{, ResNet-110 and ResNet-50 are utilized to demonstrate the efficacy of the proposed pruning method over MNIST, CIFAR-10 and ImageNet datasets. The proposed method shows the state-of-the-art pruning results for LeNet-5, VGG-16, ResNet-56, ResNet-110 and ResNet-50 architectures. In the experiments, we prune 97.98 \%, 84.85 \%, 76.89\%, 76.95\%, and 63.99\% of Floating Point Operation (FLOP)s from LeNet-5, VGG-16, ResNet-56, ResNet-110, and ResNet-50 respectively.} The proposed HRel pruning method outperforms recent state-of-the-art filter pruning methods. Even after pruning the filters from convolutional layers of LeNet-5 drastically (\ie from 20, 50 to 2, 3, respectively), only a small accuracy drop of 0.52\% is observed. Notably, for VGG-16, 94.98\% parameters are reduced, only with a drop of 0.36\% in top-1 accuracy. \textcolor{myblue}{ResNet-50 has shown a 1.17\% drop in the top-5 accuracy after pruning 66.42\% of the FLOPs.} In addition to pruning, the Information Plane dynamics of Information Bottleneck theory is analyzed for various Convolutional Neural Network architectures with the effect of pruning. The code is available at \url{https://github.com/sarvanichinthapalli/HRel}. \\ \textcolor{red}{This paper is published by Neural Networks, Elsevier. The final paper is available at:\\ \url{https://www.sciencedirect.com/science/article/pii/S0893608021004962.}}
\end{abstract}

\end{frontmatter}
\section{Introduction}
Deep Convolutional Neural Networks (CNN) are being used to provide successful and reliable solutions in various domains \cite{amodei2016deep,cheng2015deep,fayek2017evaluating,redmon2017yolo9000,ren2015faster,wang2017accurate}. In the applications of deep neural networks, the requirement for higher memory and power consumption hinders their deployment on low-end devices such as mobiles, drones. Hence, it is necessary to decrease energy consumption and memory footprint. To solve it, two types of methods have been found in literature, namely network compression and Neural Architecture Search (NAS).

Network compression is an area that accelerates the inference by reducing the Floating Point Operations (FLOPs) and decreases the memory requirement by pruning trainable parameters using various techniques. Network compression can be performed by different techniques such as \textit{network quantization, knowledge distillation, low-rank factorization} and \textit{network pruning.} \textit{Network quantization} reduces the number of bits required to represent the weights \cite{Wu2016QuantizedCN}. Binarization is an extreme case of this, where only 1 bit is used for representing weights \cite{Courbariaux2015BinaryConnectTD,10.5555/2969442.2969588}. In \textit{knowledge distillation} methods, a larger teacher model transfers its knowledge to a computationally less expensive student model \cite{44873,Romero2015FitNetsHF}. \textit{Low-rank factorization} methods aim at reducing the computational requirement by representing the convolution weight matrix as a product of low-rank matrices \cite{journals/corr/JaderbergVZ14}. \textit{Network (Parameter) pruning} methods, prune the filters in two different ways, \ie weight pruning \cite{han2015learning,lecun1990optimal} and filter pruning \cite{lin2020hrank,singh2020acceleration}. In weight pruning, the least important weights across the network are pruned. Therefore, only a few weights of the filters are pruned by weight pruning methods. Special hardware libraries are required to accelerate the network compressed by the weight pruning method. On the other hand, filter pruning methods prune the complete filter and do not require the support of any special hardware and libraries. Hence, they are widely used in the research community in recent years.

NAS based compression techniques \cite{dong2019network, guo2020dmcp,lin2020channel,liu2019metapruning, su2020locally, yu2019autoslim} are focused on finding the compact structure of neural network architecture, rather than using a criteria for computing the importance of convolutional filters. NAS methods include channel configuration \ie number of channels in each layer into the search space. Thereby the best channel configuration under various computational budgets (eg. FLOPS) is selected with less human interference.


This paper is focused on filter pruning methods which are broadly classified into two types, namely, \textit{Data free} - which use the weight matrices of filters \cite{ayinde2019redundant,basha2021deep, he2020learning, he2019asymptotic, he2018soft,he2019filter,li2016pruning, liu2017learning, singh2020leveraging,singh2020acceleration,wang2019cop,wen2020structured} and \textit{Data driven} - which use the activation maps generated by the respective filters \cite{amjad2021understanding,dai2018compressing,ganesh2020mint,hu2016network,jordao2020deep,lee2020channel,lin2020hrank,luo2017thinet,min20182pfpce}. A primitive data free filter pruning approach is proposed by Li \etal \cite{li2016pruning} that uses the filter's $\ell_1$ norm to determine the significance of filters. The filters with the least $\ell_1$ norm are considered to be less important and pruned from the model.\textcolor{myblue}{ The correlation measure between the filters is used to identify and prune the redundant filters \cite{singh2020leveraging,wang2019cop}.} Singh \etal \cite{singh2020acceleration} employed a custom regularizer based on an orthogonality constraint such that the remaining filters after pruning based on $\ell_1$ norm were independent and designed a mechanism to transfer the knowledge from the filters to be pruned to the remaining filters. Ayinde \etal \cite{ayinde2019redundant} pruned the redundant filters based on the relative cosine distance among the filters. In the \textit{data driven} category, Hu \etal \cite{hu2016network} proposed a filter pruning method that prunes filters having a greater average number of zeroes in their activations. Lin \etal \cite{lin2020hrank} pruned filters based on the rank of their respective activation maps. \textcolor{myorange}{The rank of a matrix gives the maximum number of linearly independent column vectors. Both \cite{hu2016network,lin2020hrank} have identified the significance of filters directly from their activation maps without considering the class labels (ground truths).} \textcolor{myblue}{Jordao \etal \cite{jordao2020deep} pruned filters by considering the linear relationship between filters and class labels. 
In data-free methods, it is difficult to capture the amount of relevant information retrieved by a filter about the class labels. Only in data-driven methods, the relation between transformed input (the input after applying non-linear transformations) at each hidden layer and the ground truth can be captured by an information theoretic quantity called Mutual Information (MI). MI can capture both linear and non-linear relationships. The pruning techniques \cite{amjad2021understanding,dai2018compressing,ganesh2020mint,lee2020channel} which use Mutual Information for their filter pruning criteria are data-driven methods.}

\begin{figure}[!t]
  \centering
  \includegraphics[width=0.4\textwidth]{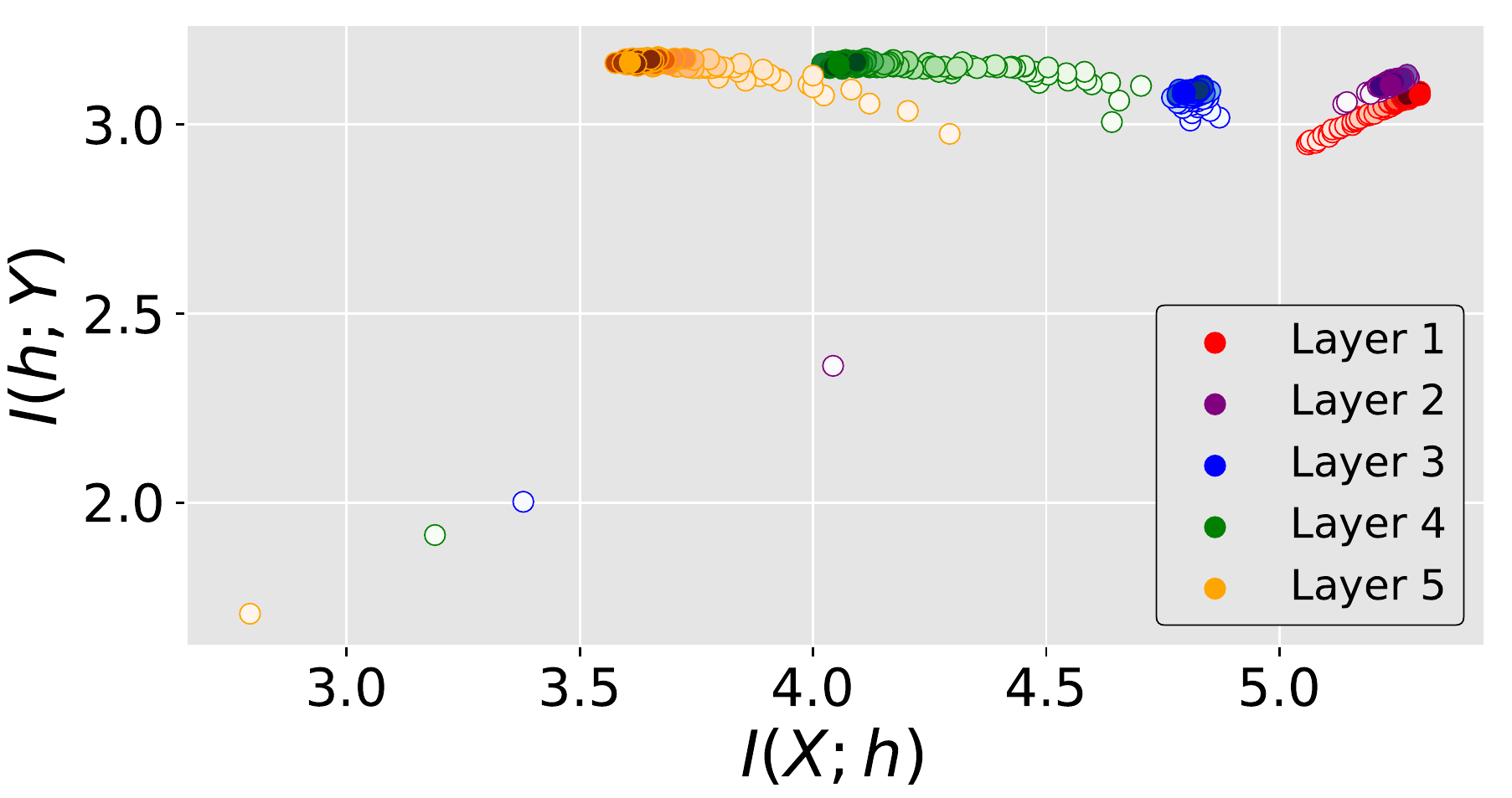} 
  \caption{Information Plane dynamics of LeNet-5 architecture trained on MNIST dataset for 20 epochs. The layers are represented with different colors and saturation of each color indicates the progress of training.}
  \label{fig:ip_dynamics}
  \end{figure}

Using Mutual Information measure, Tishby \etal proposed Information Bottleneck (IB) theory \cite{tishby2000information} and applied it in the context of neural networks \cite{shwartz2017opening}. 
 IB theory analyzes the learning of a neural network using the network's Information Plane (each hidden layer's MI with input and true class labels plotted on X-axis and Y-axis respectively) during the training as shown in Fig. \ref{fig:ip_dynamics}. From the Information Plane (IP) dynamics of IB theory, each hidden layer's MI with input and class labels increase gradually and saturate during the training.

\begin{figure}[t!]
    \centering
    \includegraphics[width=0.4\textwidth]{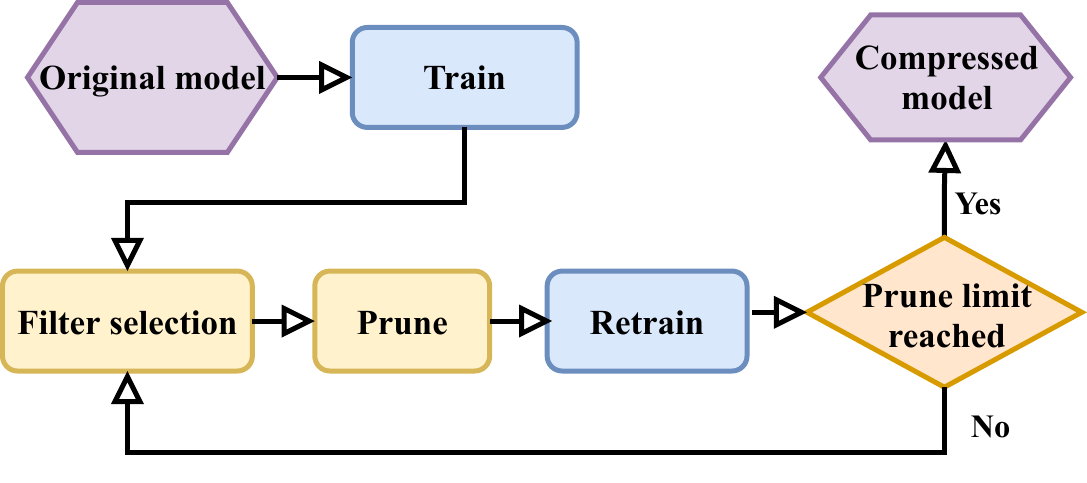}
    \caption{Illustration of complete pruning process: The original heavy model is initially trained before pruning. The pruning starts by selecting filters that are least important, followed by a retrain step. The process of pruning and retraining continues till the desired pruning limit is achieved.}
    \label{fig:process}
\end{figure}

This paper proposes a data-driven filter pruning technique for neural network classifiers based on IB theory which defines the significance of filters using their Relevance. The filters with the least Relevance from each layer are pruned iteratively. The overall pruning process is depicted in Fig. \ref{fig:process}. Among the other existing methods that use MI for defining filter's importance, only the method \cite{amjad2021understanding} observes the significance of individual neurons purely based on their Relevance. However, only fully connected and smaller neural networks are pruned by this strategy. Contrary to the method \cite{amjad2021understanding}, our method can prune filters of convolutional layers in deeper architectures like VGG-16, ResNet-56, \textcolor{myblue}{ResNet-110 and ResNet-50}. The proposed method also relies on a non-parametric estimator \cite{wickstrom2019information} that is more stable than the binning method used in \cite{amjad2021understanding} for the estimation of MI. 


\noindent Our contributions are summarized as follows:
\begin{itemize}
    \item Based on the IB theory, MI between filter's activation maps and class labels is proposed as the criterion to decide the filter's importance.
    \item The Information Plane dynamics of IB theory is shown along with the effect of pruning, which justifies the proposed filter selection criterion for pruning. 
    \item Extensive experiments show the efficacy of the proposed approach, with a considerable improvement over the recent \textcolor{myblue}{state-of-the-art methods \cite{ayinde2019redundant,basha2021deep,dai2018compressing,dong2019network,ganesh2020mint,he2020learning,he2019asymptotic,he2018soft,jordao2020deep,lee2020channel,li2016pruning,lin2020hrank,lin2020channel,lin2019towards,liu2019metapruning,singh2020leveraging,singh2020acceleration, su2020locally}.}
\end{itemize}

The rest of the article is organized as follows- Section \ref{sec:related_works} reviews the background of IB theory and MI estimation by investigating the limitations in the prior art. Section \ref{sec:proposed_method} discusses proposed HRel method. Section \ref{sec:experiments} illustrates the experimental results and Section \ref{sec:conclusion} concludes with the future directions.

\section{Related Works}
\label{sec:related_works}
This section discusses the significance of Information Plane (IP) dynamics in IB theory, limitations of the methods for MI estimation, and the limitations of the existing works that used MI for pruning.

In IB theory \cite{shwartz2017opening}, the learning process of neural networks is analyzed using the IP dynamics. During the training of neural networks, two quantities, MI of every hidden layer $h$ with input $X$ represented as $I(X;h)$ and MI of every hidden layer $h$ with label $Y$ represented as $I(h;Y)$, keep increasing. At a point during training, the quantity $I(X;h)$ starts decreasing, while $I(h;Y)$ continues to increase as shown in Fig. \ref{fig:ip_dynamics}. This is called as compression phase \cite{shwartz2017opening}. 
However, both the quantities settle at a value and do not change on further training of the neural network. There are conflicting views \cite{balda2018information,goldfeld2018estimating,saxe2019information} and supporting views \cite{chelombiev2019adaptive,jonsson2020convergence,noshad2019scalable} regarding the existence of the compression phase. The proposed method selects filters using MI between their activation maps and class labels for pruning based on IB theory.

MI estimation plays an important role in IB theory, and several works based on IB theory \cite{belghazi2018mutual,goldfeld2018estimating,saxe2019information,shwartz2017opening} use different MI estimation methods. MI calculation in deep neural networks requires the joint and marginal probabilities of high dimensional variables, which are difficult to compute. Hence various non-parametric estimators \cite{belghazi2018mutual,kolchinsky2017estimating,kraskov2004estimating,leonenko2008class,noshad2019scalable,wickstrom2019information,yasaei2019geometric} have been proposed. The basic method uses binning \cite{purwani2017analyzing} to estimate MI, where the neurons' outputs are discretized. However, the binning estimate highly depends on the bin size. The non-parametric estimators based on K-Nearest Neighbours \cite{kraskov2004estimating}, and kernel density estimation \cite{kolchinsky2017estimating,leonenko2008class} were being widely used before Mutual Information Neural Estimation (MINE) \cite{belghazi2018mutual}, which solved the problem of scaling with the sample size and dimension. An R\'{e}nyi's alpha entropy estimator \cite{wickstrom2019information} has been proposed using the matrices or tensors, which are basic entities in deep learning. It has also shown the IP dynamics on larger architecture (from the perspective of MI estimation) like VGG-16. The proposed HRel method uses matrix based R\'{e}nyi's alpha entropy estimator \cite{wickstrom2019information} for MI estimation.

In MI based filter pruning methods, Dai \etal \cite{dai2018compressing} used an upper bound of Relevance as a part of the loss function. With this modified loss function, MI between every hidden layer and the corresponding class labels increases, and MI between the consecutive layers decreases. Ganesh \etal \cite{ganesh2020mint} pruned the filters in a hidden layer with lower Mutual Information with all the other filters of the subsequent layer. Amjad \etal \cite{amjad2021understanding} have shown that in fully connected neural networks, MI between neurons in hidden layers and the corresponding class labels is a good selector for layer-wise neuron importance. Min \etal \cite{min20182pfpce} used the entropy of activations conditioned on the loss as a criterion for filter's significance. The filters with higher conditional entropy, which implies a lower MI, are pruned. Recently, Lee \etal \cite{lee2020channel} utilized gradients of MI between the activation maps of BatchNorm layers and final score vectors to the scaling factor of Batch-Normalization during back propagation to decide the filter's importance. The network architecture is augmented with an MI-subnet, which is responsible for the MI estimation.

\textcolor{mymagenta}{Compared to the existing methods, the proposed HRel method captures MI between filters and class labels (\ie Relevance) using a matrix based estimator \cite{wickstrom2019information} and uses it for filter pruning criterion. To the best of our knowledge, effect of pruning on Information plane of various CNNs is analyzed for the first time. Also, the HRel method is not employing additional architecture or changes in the loss function, unlike the MI based filter pruning methods \cite{dai2018compressing, lee2020channel}.  }

\section{Proposed HRel Pruning Approach}
\label{sec:proposed_method}
In this section, we propose an HRel filter pruning approach for convolutional neural networks. The filters are pruned depending on their Relevance in corresponding hidden layers. The Relevance criterion is chosen based on the IB theory using the Mutual Information metric. This section describes the basic definitions and notations, computation of the Relevance followed by steps of filter pruning.
\begin{figure}[t!]
    \centering
    \includegraphics[width=0.5\textwidth]{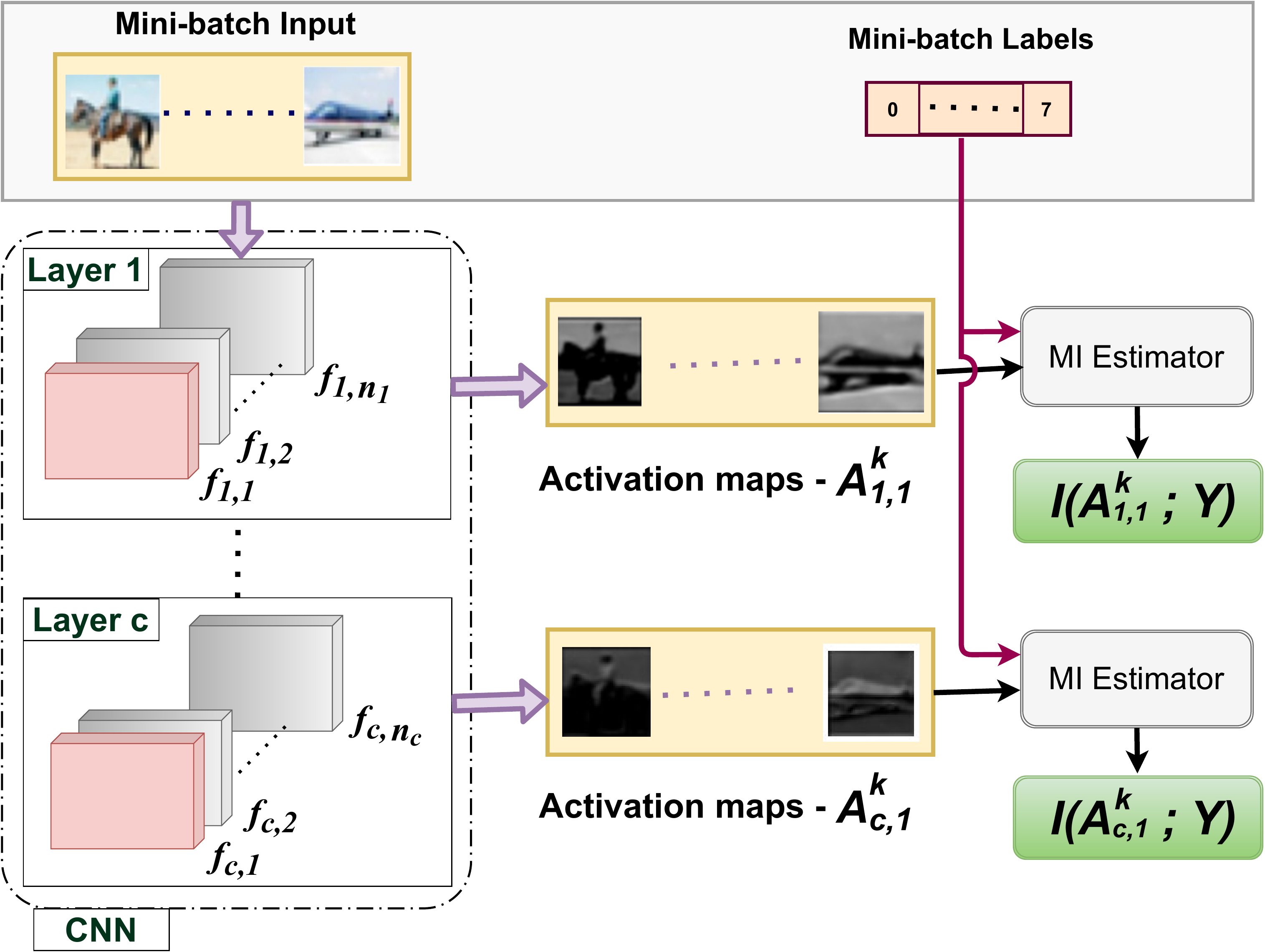}
    \caption{The steps involved in calculating the Relevance of $A_{i,j}^k$, \ie activation map of the $j^{th}$ filter of $i^{th}$ layer from the $k^{th}$ mini-batch.}
    \label{fig:method}
\end{figure}
\subsection{Basic Definitions and Notations} 
Mutual Information (MI) between two random variables U, V i.e., $ \textcolor{mymagenta}{I\mathrm {(U;V)}}$ quantifies the amount of information that can be inferred about a random variable U by observing the other random variable V or vice versa, which is expressed as
\begin{equation}
\label{eq:MI}
    \textcolor{mymagenta}{I\mathrm {(U;V)= H(U)+ H(V) - H(U,V)}}
\end{equation}

where $ \textcolor{mymagenta}{\mathrm {H(U)}}$ and $ \textcolor{mymagenta}{\mathrm {H(V)}}$ denote \textit{entropy} \cite{10.5555/1146355}, $ \textcolor{mymagenta}{\mathrm {H(U,V)}}$ denotes \textit{joint entropy} \cite{10.5555/1146355}.

Assume a CNN model having $c$ convolutional layers, in which $L_i$ is the $i^{th}$ convolutional layer. The filters of a convolutional layer $L_i$ can be represented as $\mathcal{F}_{L_{i}}$ = \{$f_{i,1}, f_{i,2}, ...., f_{i,n_i}$\}, where $n_i$ is the number of filters in layer $L_i$, $f_{i,j} \in \mathbb{R}^{n_{i-1} \times d_i \times d_i}$, $d_i$ is the kernel size and \textcolor{mymagenta}{ $n_{i-1}$ is the number of channels in each filter which is same as the depth of the activation input}. Suppose there are $m$ mini-batches of input for training the network. For the $k^{th}$ mini-batch, the activation maps of filters from $i^{th}$ hidden layer is denoted by $\mathcal{A}_i^k$= \{$A_{i,1}^k$, $A_{i,2}^k$,.... $A_{i,n_i}^k$\} $\in$ $\mathbb{R}^{n_i \times s \times h_i \times w_i}$, where $n_i$ is the number of filters, $s$ is the mini-batch size, $h_i$ and $w_i$ are the height and width of the activation maps, respectively. Notably, $A_{j,1}^k$ $\in$ $\mathbb{R}^{s \times h_i \times w_i}$ is the activation map generated by $f_{i,j}$ for all the samples in $k^{th}$ mini-batch. During pruning, filters in layer $L_i$ are split into Pruned filters $\mathcal{P}_{L_i} =\{ f_{i,P_{1}},f_{i,P_{2}},....,f_{i,P_{p_i} }\}$ and Remaining filters $\mathcal{R}_{L_i}=\{ f_{i,R_{1}},f_{i,R_{2}},....,f_{i, R_{r_i}}\}$, where $p_i$ and $r_i$ are the number of pruned and remaining filters of layer $L_i$. $P_{j}$ and $R_{k}$ denote $j^{th}$, $k^{th}$ filters in pruned and remaining filter set, respectively. Notably, $\mathcal{P}_{L_i} \cap \mathcal{R}_{L_i}= \varnothing $, $\mathcal{P}_{L_i} \cup \mathcal{R}_{L_i}= \mathcal{F}_{L_i} $, and $p_i + r_i = n_i$.

\begin{figure*}[t!]

     \centering
     \begin{subfigure}[b]{0.4\columnwidth}
         \centering
         \includegraphics[width=\textwidth]{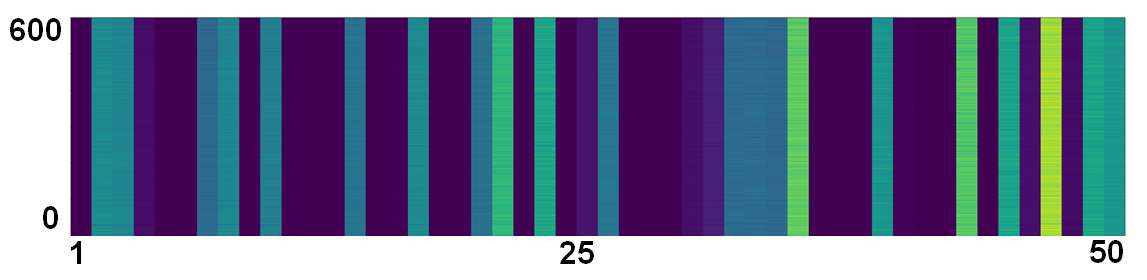}
         \caption{LeNet-5: 50 Filters}
         \label{fig:y equals x}
     \end{subfigure}
     \hfill
     \begin{subfigure}[b]{0.4\columnwidth}
         \centering
         \includegraphics[width=\textwidth]{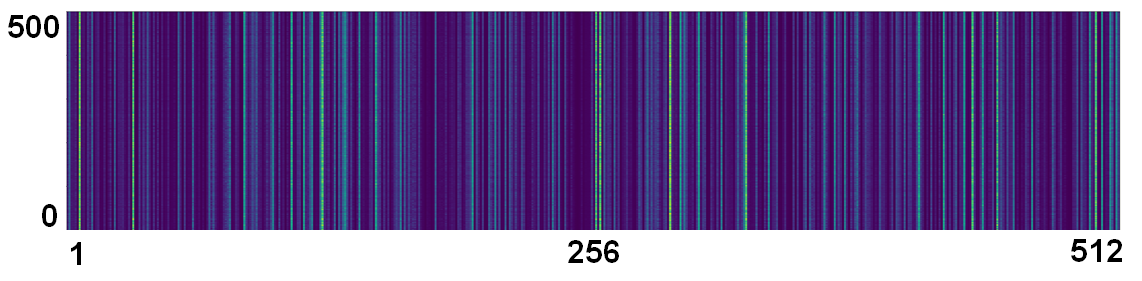}
         \caption{VGG-16: 512 Filters}
         \label{fig:three sin x}
     \end{subfigure}
     \hfill
     \begin{subfigure}[b]{0.4\columnwidth}
         \centering
         \includegraphics[width=\textwidth]{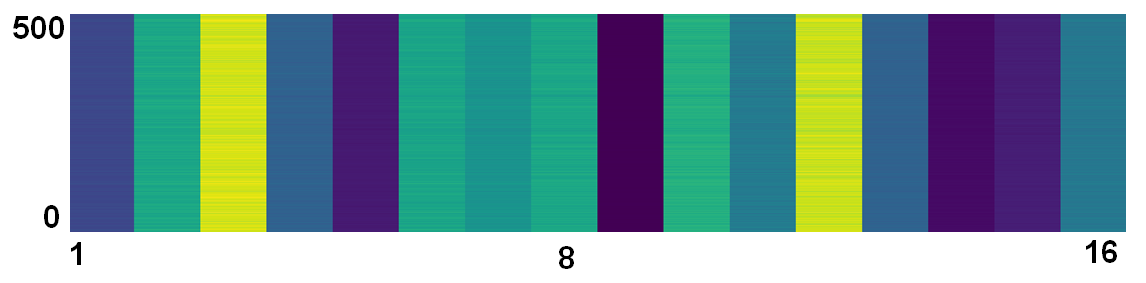}
         \caption{ResNet-56: 16 Filters}
         \label{fig:five over x}
     \end{subfigure}
     \hfill
     \begin{subfigure}[b]{0.4\columnwidth}
         \centering
         \includegraphics[width=\textwidth]{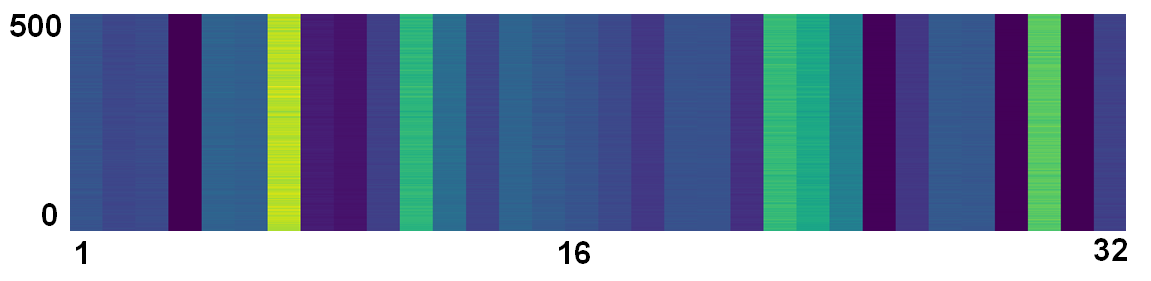}
         \caption{ResNet-110: 32 Filters}
         \label{fig:five over x}
     \end{subfigure}
     \hfill
     \begin{subfigure}[b]{0.4\columnwidth}
         \centering
         \includegraphics[width=\textwidth]{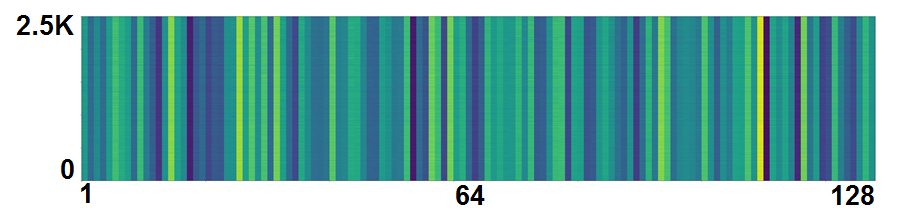}
         \caption{ResNet-50: 128 Filters}
         \label{fig:five over x}
     \end{subfigure}

     \begin{subfigure}[b]{0.4\columnwidth}
         \centering
         \includegraphics[width=\textwidth]{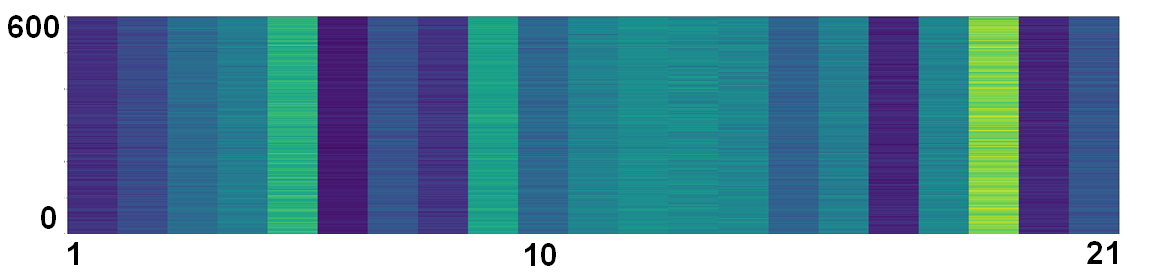}
         \caption{LeNet-5: 21 Filters}
         \label{fig:y equals x}
     \end{subfigure}
     \hfill
     \begin{subfigure}[b]{0.4\columnwidth}
         \centering
         \includegraphics[width=\textwidth]{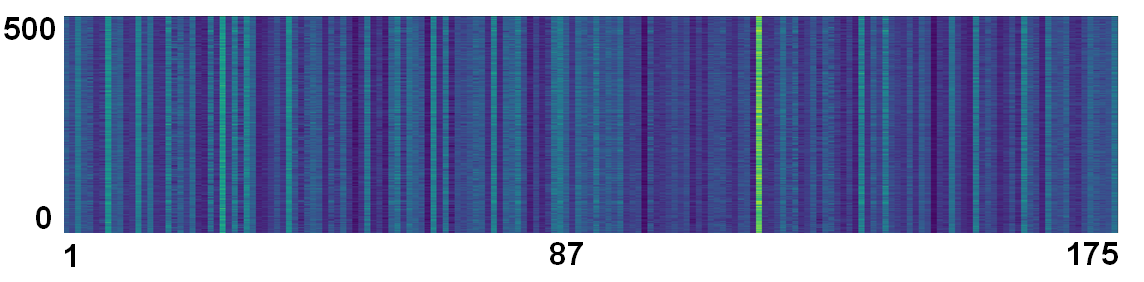}
         \caption{VGG-16: 175 Filters}
         \label{fig:three sin x}
     \end{subfigure}
     \hfill
     \begin{subfigure}[b]{0.4\columnwidth}
         \centering
         \includegraphics[width=\textwidth]{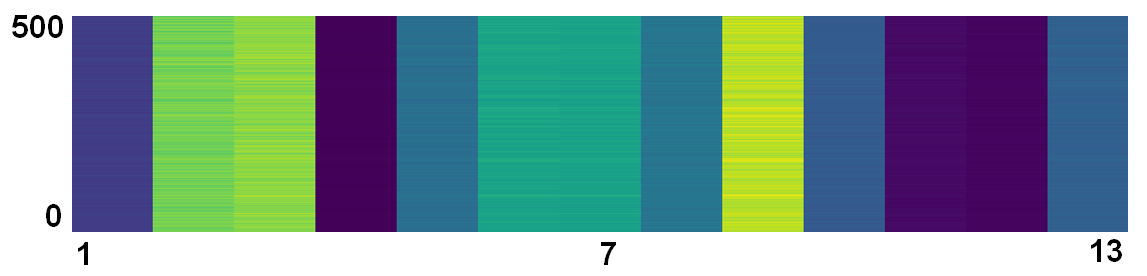}
         \caption{ResNet-56: 13 Filters}
         \label{fig:five over x}
     \end{subfigure}
     \hfill
     \begin{subfigure}[b]{0.4\columnwidth}
         \centering
         \includegraphics[width=\textwidth]{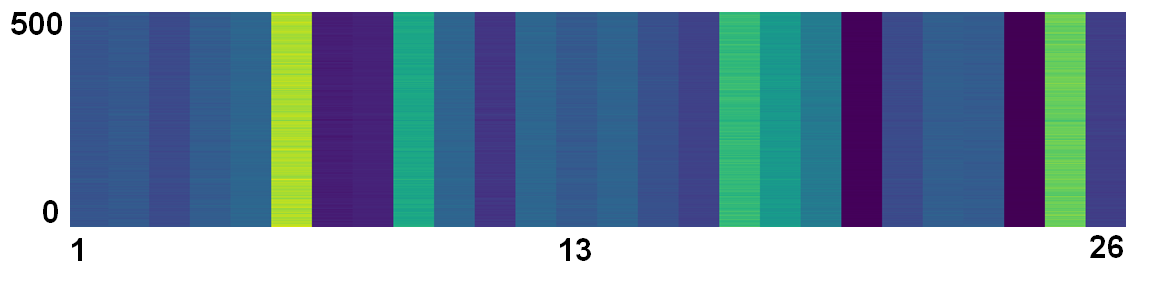}
         \caption{ResNet-110: 26 Filters}
         \label{fig:five over x}
     \end{subfigure}
     \hfill
     \begin{subfigure}[b]{0.4\columnwidth}
         \centering
         \includegraphics[width=\textwidth]{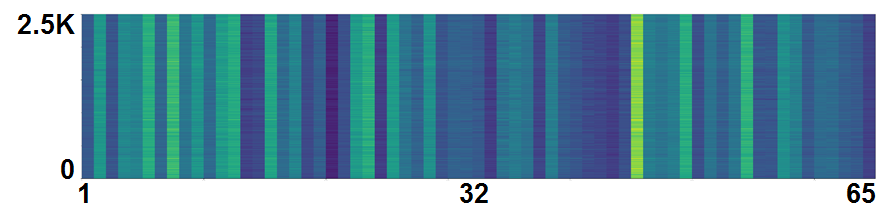}
         \caption{ResNet-50: 80 Filters}
         \label{fig:five over x}
     \end{subfigure}


     \begin{subfigure}[b]{0.4\columnwidth}
         \centering
         \includegraphics[width=\textwidth]{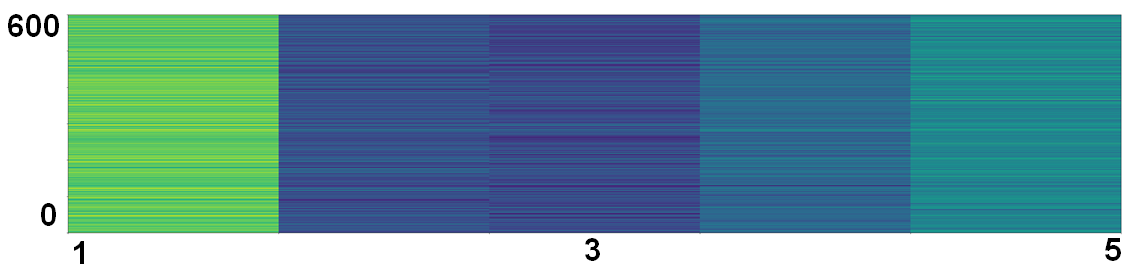}
         \caption{LeNet-5: 5 Filters}
         \label{fig:y equals x}
     \end{subfigure}
     \hfill
     \begin{subfigure}[b]{0.4\columnwidth}
         \centering
         \includegraphics[width=\textwidth]{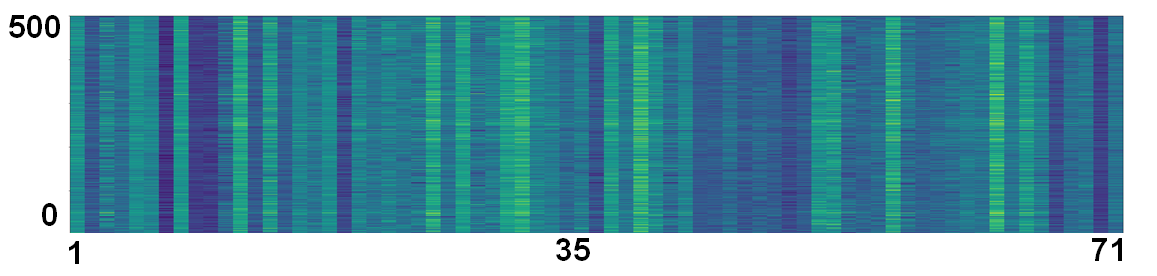}
         \caption{VGG-16: 71 Filters}
         \label{fig:three sin x}
     \end{subfigure}
     \hfill
     \begin{subfigure}[b]{0.4\columnwidth}
         \centering
         \includegraphics[width=\textwidth]{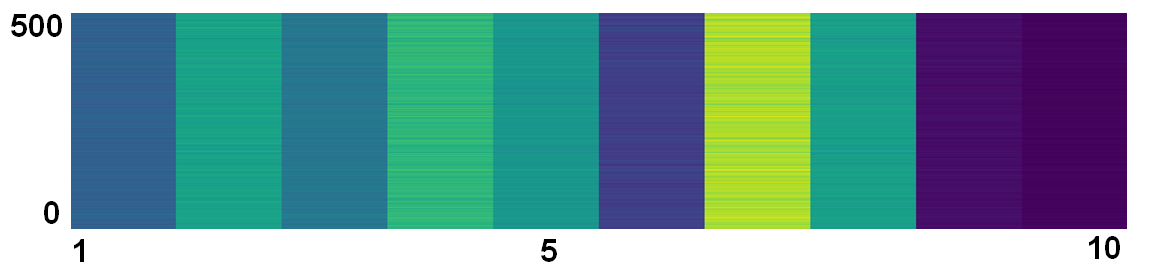}
         \caption{ResNet-56: 10 Filters}
         \label{fig:five over x}
     \end{subfigure}
     \hfill
     \begin{subfigure}[b]{0.4\columnwidth}
         \centering
         \includegraphics[width=\textwidth]{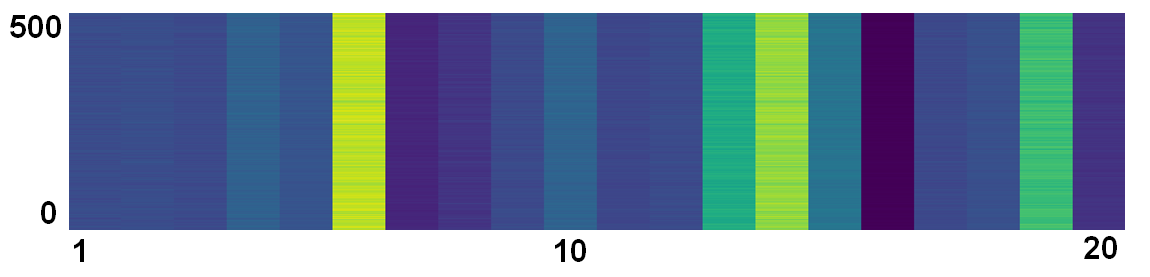}
         \caption{ResNet-110: 20 Filters}
         \label{fig:five over x}
     \end{subfigure}
     \hfill
     \begin{subfigure}[b]{0.4\columnwidth}
         \centering
         \includegraphics[width=\textwidth]{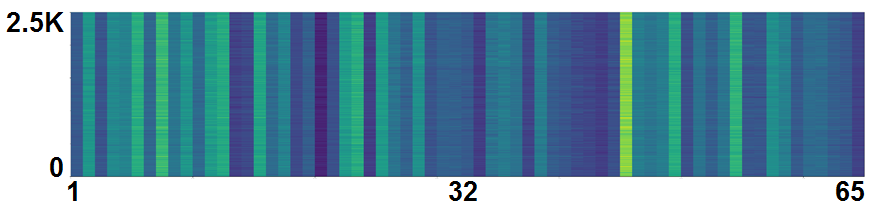}
         \caption{ResNet-50: 65 Filters}
         \label{fig:five over x}
     \end{subfigure}
     \\
     \begin{subfigure}[b]{0.7\columnwidth}
         \centering
         \includegraphics[width=\textwidth]{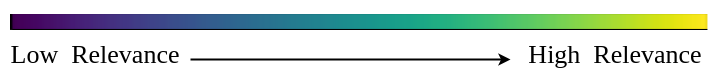}
         \label{fig:five over x}
     \end{subfigure}
        \caption{The Relevance of the remaining filters from convolution layers of different architectures (the number of remaining filters are specified for each sub-figure). For each sub-figure, X-axis denotes all the remaining filters in a convolutional layer at different pruning iterations. Y-axis (bottom to top) denotes the mini-batches of the training data. The first row depicts the architectures before pruning. Rows 2, 3 indicate the Relevance of filters during specific pruning iterations. Convolutional layers 2, 9, 13, 37 and 34 of LeNet-5, VGG-16, ResNet-56, ResNet-110, and ResNet-50 respectively are used.}
        \label{fig:mi_over_epoch}
\end{figure*}
\subsection{The Relevance as Filter Selection Criteria}
\label{subsec:rel_computation}
The proposed HRel pruning method utilizes the Relevance, which is a key component of Information Plane dynamics used in IB theory. Though there is an ongoing debate on the existence of the compression phase, there is no ambiguity in the increment and saturation of hidden layers' Relevance ($I(L_i;Y)$) \ie MI between each hidden layer's ($L_i$) activation maps and the class labels ($Y$) in neural networks, during training. It is also observed that initially, all the layers have less Relevance at the beginning of training. But as the training progresses, the Relevance of each layer also gradually increases and gets saturated, as shown in Fig. \ref{fig:ip_dynamics}. So, a higher Relevance gained by the hidden layers during the training implies that the hidden layers learned more relevant information about the class labels. Similar to hidden layers, individual filter's Relevance ($I(f_{i,j};Y)$) \ie MI between each filter's activation maps and the class labels, also determines the amount of relevant information extracted by filter about the class labels. Hence, the Relevance of filters is employed in the proposed method to determine the significance of the filters across each layer. For $k^{th}$ mini-batch of training data, the Relevance between the activation maps of filters and class labels $Y$ given by $I(A_{i,j}^k;Y)$ is obtained as shown in Fig. \ref{fig:method}.

 \begin{algorithm}[t!]
	\caption{\textbf{:} HRel pruning of a layer $L_i$ for a pruning iteration} 
	\label{algo:filter_selection}
	\bigskip
	\textbf{Inputs:} $\mathcal{R}_{L_i}$- Set of remaining filters in $L_i$, $prune\_ratio_i$ - The \begin{addmargin}[1em]{0em}percentage of filters to prune in each pruning iteration, \textcolor{mymagenta}{$r_i$ - The number of remaining filters of layer $L_i$,} $limit_i$-Number of filters to be retained in layer $L_i$ of final compressed model\end{addmargin}
	\smallskip
	\textbf{Output:} Updated $\mathcal{R}_{L_i}$
    \smallskip
	\begin{algorithmic}[1]
	    \color{mymagenta}
	    \If {first \textit{pruning iteration}}
	        \State $\mathcal{R}_{L_i}\leftarrow$ $\mathcal{F}_{L_{i}}$
	
	        $r_i \leftarrow n_i$ 
	    
	    \EndIf
	    \color{black}
	    \If {$r_i > limit_i$}

	        \For {\textcolor{mymagenta}{each mini-batch $k$ of total m mini-batches,} $k$ $\in $ 1,2,...,m}
	            \For {each filter $f_{i,j}  \in  \mathcal{R}_{L_i}$}
	                \State compute $I(A_{i,j}^k;Y)$ using subsection \ref{subsec:rel_computation}
	            \EndFor
	        \EndFor
	        \For {Each filter $f_{i,j}  \in  \mathcal{R}_{L_i}$ }
	            \State $ I(f_{i,j};Y) \leftarrow \frac{\sum_{k=1}^{m} I(A_{i,j}^k;Y)}{m}$
	        \EndFor
	    
	    \State $t_i \leftarrow \lceil r_i \times prune\_ratio_i / 100  \rceil$
	    \State Sort $\mathcal{R}_{L_i}$ in ascending order ;
	    \State
	    $\mathcal{R}^{sorted}_{L_i}\leftarrow \{ f_{i,\textcolor{mymagenta}{R_1}},f_{i,\textcolor{mymagenta}{R_2}},....,f_{i,\textcolor{mymagenta}{R_{r_i}}}\}; \textcolor{mymagenta}{R_1,R_2,...R_{r_i}}$ is a permutation of filters in $\mathcal{R}_{L_i}$ :
	    $I(f_{i,\textcolor{mymagenta}{R_1}};Y) \leq I(f_{i,\textcolor{mymagenta}{R_2}};Y) \leq...\leq I(f_{i,\textcolor{mymagenta}{R_{t_i}}};Y)\leq...\leq I(f_{i,\textcolor{mymagenta}{R_{r_i}}};Y)$
	    \State $\mathcal{R}^{prune}_{L_i}\leftarrow\{ f_{i,\textcolor{mymagenta}{R_1}},f_{i,\textcolor{mymagenta}{R_2}},....,f_{i,\textcolor{mymagenta}{R_{t_i}}}\}$
        \State $\mathcal{R}_{L_i}\leftarrow \mathcal{R}_{L_i}- \mathcal{R}^{prune}_{L_i}$
        \State $r_i\leftarrow r_i - t_i$
	    \EndIf
	\end{algorithmic} 
	
\end{algorithm}
\vspace{-1mm}

The proposed method estimates two (Relevance) quantities $I(L_i;Y)$ and $I(f_{i,j};Y)$, which are utilized in IP dynamics and filter pruning, respectively. The computation of $I(f_{i,j};Y)$ uses the activation maps generated by each filter from a hidden layer, whereas computation of $I(L_i;Y)$ uses the complete output of a hidden layer. Estimation of MI between the activation maps and class labels in the proposed method is similar to \cite{wickstrom2019information}.

For a given mini-batch of size \textcolor{mymagenta}{s} with the activation maps generated $X =\{x_1,x_2...,\textcolor{mymagenta}{x_s}\}$, the Gram matrix \textcolor{mymagenta}{G} of size \textcolor{mymagenta}{$s \times s $} is calculated using Gaussian kernel as $\mathrm {\textcolor{mymagenta}{G_{i,j}}}= e^{ -\ \frac{1}{\sigma ^2} \parallel x_i - x_j \parallel ^2_F}$ where $\mathrm {\textcolor{mymagenta}{G}} \in \mathbb{R}^{\textcolor{mymagenta}{s\times s}}$ for all $ {\textcolor{mymagenta}{\mathrm i,j \in [1, s]}}$,
 $ \sigma $ denotes kernel width, and $\parallel . \parallel_F$ denotes Frobenius Norm. As presented in \cite{nielsen2002quantum}, entropy is consequently calculated using the Eigen values of the normalized Gram matrix \textcolor{mymagenta}{$\mathrm {N}$} as 
\begin{equation}
\label{eq:entropy}
     \textcolor{mymagenta}{\mathrm {{H}(N)}= -\sum_{\mathrm{i=1}}^{\mathrm{m}} \lambda_i \log_2\lambda_i }
\end{equation}
\textcolor{mymagenta}{Where $\mathrm {N_{i,j}=\frac{1}{s} \frac{ G_{i,j}}{\scriptstyle \sqrt{\mathrm{ G_{i,i}G_{j,j}}}}}$, $\mathrm {N \in \mathbb{R}^{s\times s}}$ for all $\mathrm {i,j \in [1, s]},$} and $\lambda_i$ is the $\mathrm {i^{th}}$ eigen value of \textcolor{mymagenta}{$\mathrm {N}$}.

\begin{figure*}[!t]
  \begin{subfigure}{0.4\columnwidth}
  \includegraphics[width=\textwidth]{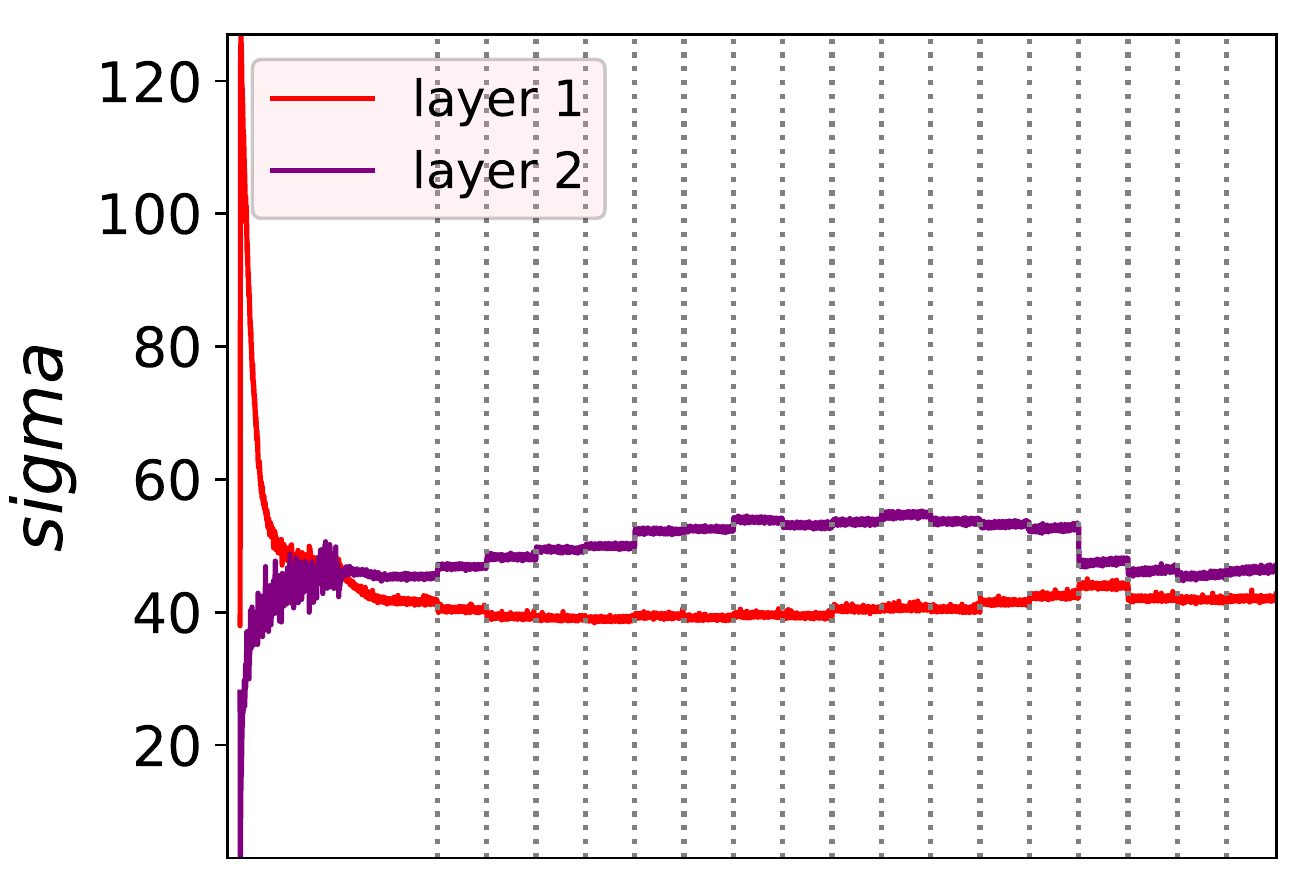}
  \caption{LeNet-5 on MNIST}
  \end{subfigure}
  \hfill
  \begin{subfigure}{0.4\columnwidth}
  \includegraphics[width=\textwidth]{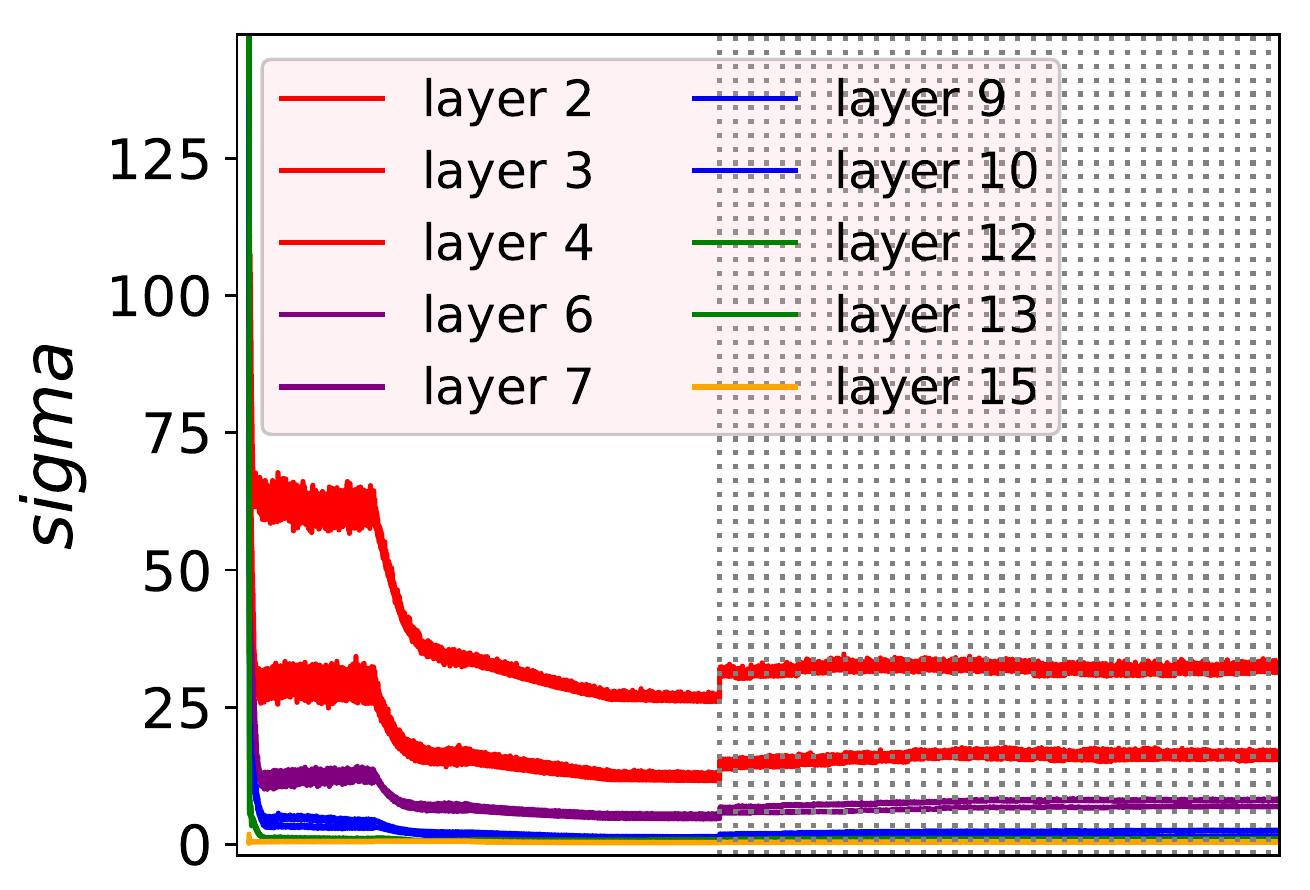} 
  \caption{VGG-16 on CIFAR10} 
  \end{subfigure} 
  \hfill
  \begin{subfigure}{0.4\columnwidth} 
  \includegraphics[width=\textwidth]{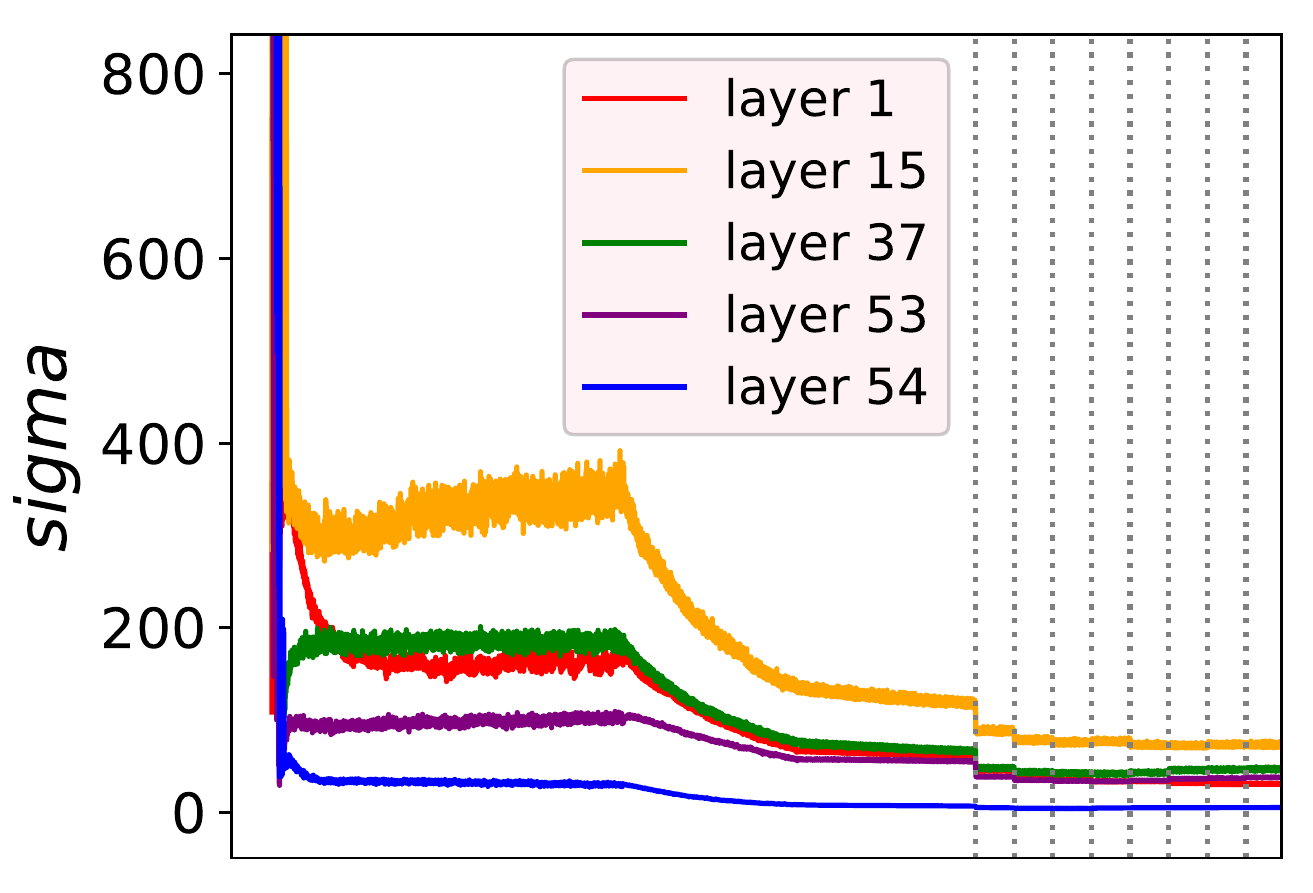} 
  \caption{ResNet-56 on CIFAR10} 
  \end{subfigure}  
  \hfill 
  \begin{subfigure}{0.4\columnwidth} 
  \includegraphics[width=\textwidth]{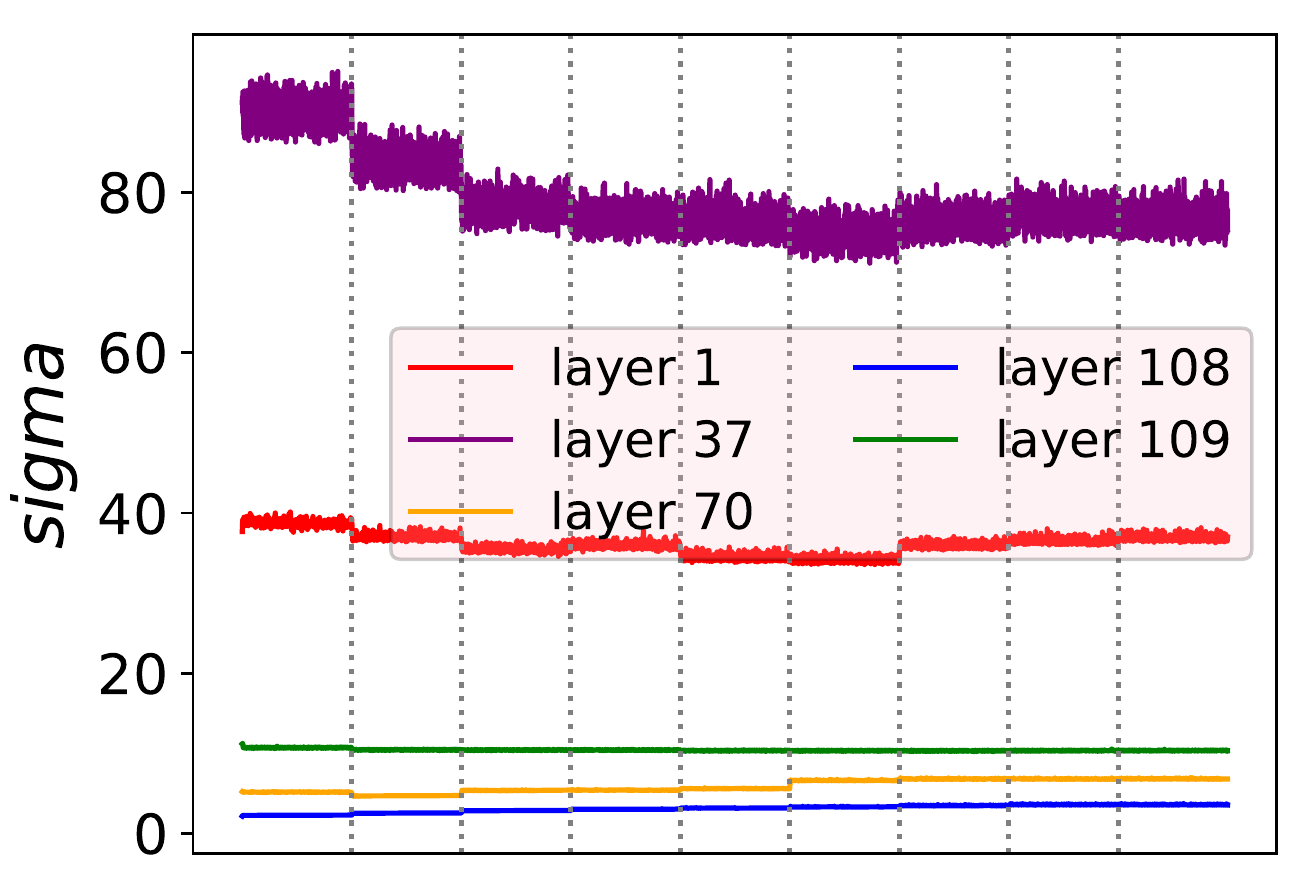} 
  \caption{Resnet-110 on CIFAR10}
  \end{subfigure}
  \hfill 
  \begin{subfigure}{0.4\columnwidth} 
  \includegraphics[width=\textwidth]{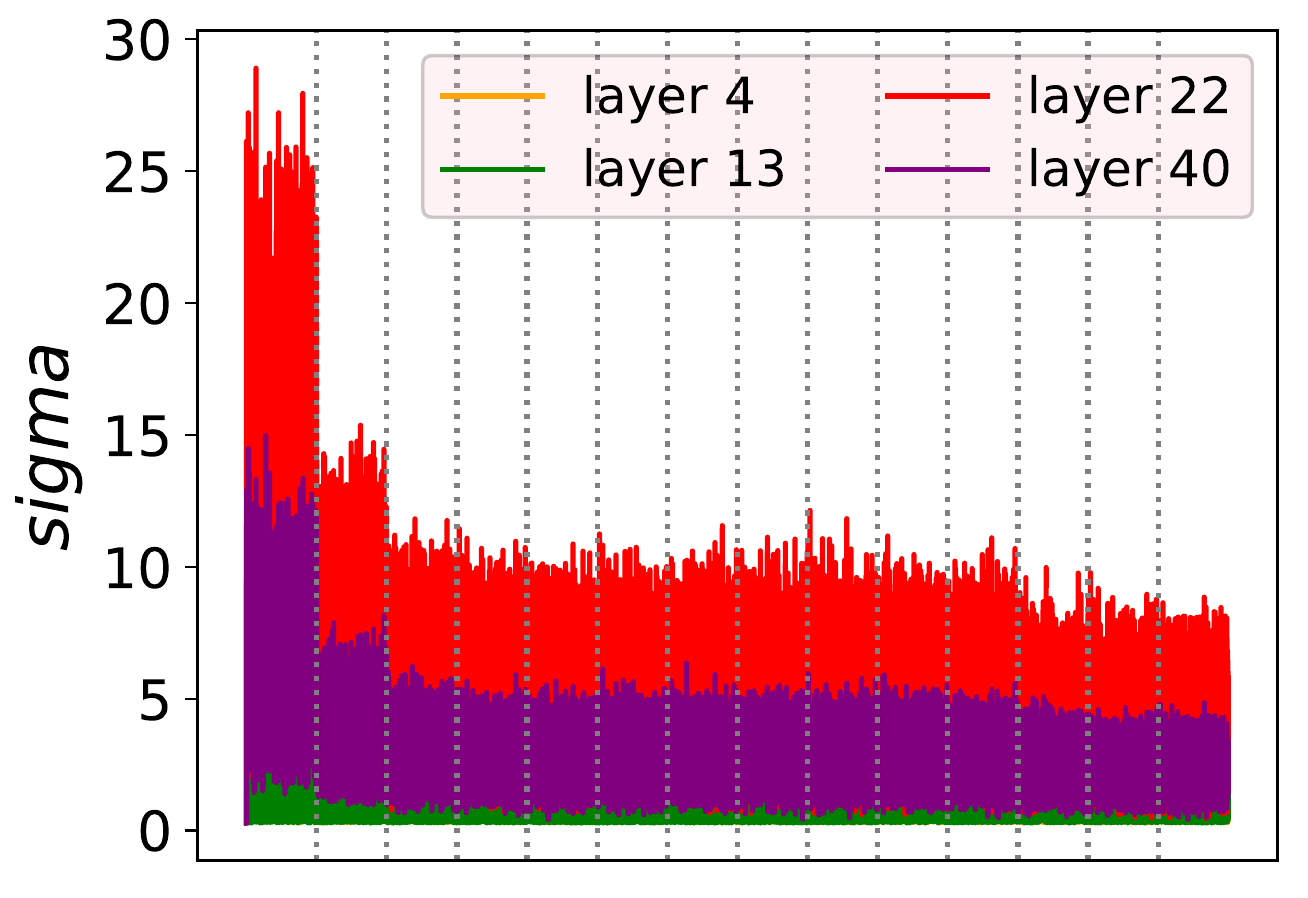} 
  \caption{Resnet-50 on ImageNet}
  \end{subfigure}

  \caption{Kernel width $\sigma_i$ for different layers denoted by sigma across the pruning iterations for different \textit{architecture + dataset} combinations. The X-axis denotes training iterations from left to right (each vertical line in the sub-figures denotes a pruning iteration).}
  \label{fig:sigmas}
  \end{figure*}

 The joint entropy between random variables \textcolor{mymagenta}{U, V} is calculated using Hadamard product ($\circ$) of their normalized Gram matrices \textcolor{mymagenta}{$\mathrm {N_U}$, $\mathrm {N_V}$}, respectively \cite{giraldo2014measures} as
\begin{equation}
\label{eq:joint_entropy}
    \textcolor{mymagenta}{\mathrm { H(U,V)= H(N_{{U}} \circ N_{{V}})} }
\end{equation}

The Relevance is calculated using the Equations \ref{eq:MI}, \ref{eq:entropy} and \ref{eq:joint_entropy}. Unlike rank, the Relevance of activation maps generated by filters changes with the mini-batches, as each filter can share a different amount of information with different classes as shown in Fig. \ref{fig:mi_over_epoch}. Slight variations in the Relevance values of filters over mini-batches in an epoch can also be observed. Therefore the mean value of the Relevance of filters measured across \textcolor{myblue}{the mini-batches} of training data is considered. \textcolor{myorange}{The averaged Relevance values across the mini batches for the filters in Fig. 4 are discussed in subsection \ref{results:distribution} and depicted in Fig. \ref{fig:Relevance_distribution}.}
It is also observed that in Fig. \ref{fig:mi_over_epoch} for each architecture except ResNet-50 from top to bottom, the color saturation of the plots gradually moved towards the brighter side.


\subsection{Filter Pruning Steps}
The pruning of filters involves three main stages. Firstly, the neural network is trained till the baseline accuracy is achieved. Secondly, each filter's Relevance is computed and filters with low Relevance are pruned. Thirdly, the network is retrained. The filter pruning and retraining are done iteratively.

\subsubsection{Initial Training}
The network parameters ($\Theta$) are initialized 
\textcolor{myblue}{and updated until the model convergence}. Training of network with each mini-batch of data is called \textit{training iteration}. During network training, the kernel width $\sigma_i$ for each hidden layer $L_i$ is computed for all mini-batches. 
\subsubsection{Filter Selection and pruning}

After initial training, pruning, and retraining of the network are performed iteratively, where each iteration is called \textit{pruning iteration}. In each pruning iteration, the filters selected using Algorithm \ref{algo:filter_selection} are pruned, and the network is retrained. To select the filters to be pruned from each layer $L_i$, two hyperparameters are required. One is the final number of filters to be retained denoted by $limit_i$, and the other is the percentage of remaining filters to be pruned at each iteration of pruning identified by $prune\_ratio_i$. The training data is processed batch-wise and the Relevance of each filter \textcolor{myblue}{in a given} mini-batch $k$ \ie $I(A_{i,j}^{k};y)$, is calculated using steps 2-6 in Algorithm \ref{algo:filter_selection}. $I(f_{i,j};y)$ is then obtained by averaging $I(A_{i,j}^{k};y)$ across \textcolor{myblue}{ the mini-batches}, for all filters in each layer as shown in steps 7-9. Next, the filters in each layer are sorted based on $I(f_{i,j};y)$ value and top $prune\_ratio_i$ \% of the filters in each layer $L_i$ given by $t_i$ are pruned. Step 11 in Algorithm \ref{algo:filter_selection} can be skipped if a fixed number of filters are removed from each layer by directly specifying $t_i$ value. The remaining filters and their count are updated using steps 14-16 in Algorithm \ref{algo:filter_selection}.

\subsubsection{Retraining}
This section focuses on improving the model performance lost due to the pruning of filters. After pruning the selected filters from the network with $c$ convolutional layers, the model contains a reduced set of the trainable parameters $\Theta^'$ compared to the original model $\Theta$.
\begin{equation}
    \Theta^' = \Theta \setminus \{\mathcal{P}_{L_i} \hspace{1mm}| \hspace{1mm} i \in 1,2, \dots,c\} 
\end{equation}
The network is then fine-tuned by retraining each architecture for a certain number of epochs to regain the accuracy drop. 

It has been \textcolor{myblue}{observed} that the kernel width $\sigma_i$ calculated for each layer along with the initial training saturates after a few epochs as shown in Fig. \ref{fig:sigmas} (i.e., before the pruning iterations begin). Therefore, its calculation is deterred in further training. In our work, we observed $\sigma_i$ values across the pruning iterations and found that despite pruning, the values did not fluctuate much, as shown in Fig. \ref{fig:sigmas}. Hence the computation of $\sigma_i$ is not done during the retraining of the network after pruning. The final obtained kernel width of hidden layers during pruning is used for both hidden layers and the corresponding filters across the pruning iterations.

\section{Experiments and Analysis}
\label{sec:experiments}
\renewcommand\thesubsection{\textit{\thesection}.\textit{\Alph{subsection}}}
To demonstrate the potency of the proposed HRel pruning method, the following dataset + architecture combinations - 
MNIST \cite{lecun-mnisthandwrittendigit-2010} + LeNet-5 \cite{lecun-mnisthandwrittendigit-2010}, CIFAR10 \cite{krizhevsky2009learning}+ VGG-16 \cite{simonyan2014very}, CIFAR10 + ResNet-56 \cite{he2016deep}, CIFAR10 + ResNet-110 \cite{he2016deep}, \textcolor{myblue}{ImageNet\cite{russakovsky2015imagenet} + ResNet-50 \cite{he2016deep}} are used. Experimental results are analyzed in terms of accuracy, IP dynamics, and the Relevance distribution. Subsections \ref{results:lenet}, \ref{results:vgg}, \ref{results:resnet56}, \ref{results:resnet110} and \ref{results:resnet50} compare the pruning results with state-of-the-art methods. Furthermore, the subsection \ref{results:ip_plane} analyzes the IP dynamics related to IB theory and finally, the subsection \ref{results:distribution} examines the distribution of the Relevance values of filters across the pruning iterations.

The percentage of filters to be removed from each layer and the number of filters to be retained in the final network are the hyperparameters for each of the architectures. The same CNNs as in the recent work based on the rank of activation maps \cite{lin2020hrank}, are utilized for verifying the efficacy of the proposed filter pruning method. \textcolor{myblue}{The batch size is modified to $80$ in ResNet-50 and $100$ in rest of the architectures. Nesterov momentum \cite{botev2017nesterov} is used for ResNet-56 and ResNet-110.} Same settings as in \cite{wickstrom2019information} are used for MI estimation by using an input kernel of width $8$ and a label kernel of width $0.1$. \textcolor{myblue}{During the calculation of filters' Relevance batch size of $128$ in ResNet-50 and $100$ in rest of the architectures are used.} The kernel width of each hidden layer for every dataset + architecture combination is evaluated until the baseline accuracy is achieved. The hyperparameters such as learning rate, learn rate schedule for training and pruning iterations are specified for each architectures in subsections \ref{results:lenet}, \ref{results:vgg}, \ref{results:resnet56}, \ref{results:resnet110} and \ref{results:resnet50}.

In our experimental setting, we have evaluated the proposed method in terms of \textit{FLOPs}- Floating Point Operations and \textit{Trainable Parameters} for all the models. For a fair comparison with the other methods, parameters and FLOPs corresponding to Convolutional and Fully Connected layers alone are considered. \textcolor{myblue}{ FLOPs and Params in the results table (Table \ref{tab:lenet_results}, \ref{tab:vgg_results}, \ref{tab:res56_results}, \ref{tab:res110_results}, and \ref{tab:res50_results}) denote remaining FLOPs and remaining parameters after pruning, respectively. M denotes Millions ($10^6$) and B denotes Billions ($10^9$) in the columns of FLOPs and Params.} The percentage of pruned FLOPs and pruned Parameters are denoted by $P_f$\% and $P_p$\%, respectively. \textcolor{mymagenta}{The baseline accuracy and the accuracy after pruning (in percentage) is denoted by Acc$_{baseline}\%$ and Acc$_{pruned}$\% respectively. Accuracy Drop is denoted by Acc$   \big\downarrow$. For ResNet-50 Top-1,Top-5 implies Top-1 and Top-5 baseline accuracies. Top-\#$_{pruned}$\% and Top-\#$\big\downarrow$ denote corresponding accuracy and accuracy drop, after pruning.}

\subsection{LeNet-5 on MNIST Dataset}
\label{results:lenet}
MNIST is a handwritten digits dataset, that contains 60,000 training images and 10,000 test images, each of size $28 \times 28 \times 1 $. LeNet-5 architecture contains 2 convolutional layers, having 20 and 50 filters with the spatial dimension of $5 \times 5$, followed by 3 fully connected layers with 800, 500, and 10 neurons, respectively. The network is trained for 40 epochs with the initial learning rate of 0.1, which is divided by 10 at epoch numbers 20 and 30 to achieve the baseline accuracy. For the proposed HRel method, rather than pruning an equal percentage of filters from each layer, better results are observed empirically if initial layers are pruned at a lower rate compared to final layers. Thus, in each pruning iteration, $4\%$ and $12\%$ of the filters are pruned from the first and second convolutional layers, respectively. After pruning, the network is retrained for 40 epochs beginning with a learning rate of 0.1, which is divided by 10 at epochs 10 and 20. LeNet-5 pruning results are compared with benchmark methods in Table \ref{tab:lenet_results}. Note that HRel-\# represents HRel at different pruning limits.

\textcolor{myblue}{HRel method achieves
a higher FLOPs reduction percentage, i.e., 97.98\%, with the accuracy of 98.78\%, and accuracy drop of 0.52, outperforming CFP \cite{singh2020leveraging} and HBFP \cite{basha2021deep} methods, for an equal FLOPs reduction percentage. While PP-OC \cite{singh2020acceleration} has the least accuracy drop, the HRel method achieved the higher test accuracy when \{20, 50\} filters are pruned to \{4, 5\} filters in the first and second convolution layers, respectively.} Though VIB \cite{dai2018compressing} method had a lesser number of remaining FLOPs, \ie $0.09$M, the authors have mentioned that they considered half the number of FLOPs. Hence, it would account for $0.18$M to compare with all the other methods. \textcolor{myblue}{Inspite of having more baseline accuracy than CFP and HBFP the accuracy drop is considerably less. }

\subsection{VGG-16 on CIFAR-10 Dataset}
\label{results:vgg}
CIFAR-10 dataset consists of 50,000 training images and 10,000 test images belonging to 10 classes. The image size is $32 \times 32 \times 3$. The proposed HRel method is applied to VGG-16 architecture 64-64-128-128-256-256-256-512-512-512-512-512-512-512-10 with 13 convolutional layers and 2 fully connected layers to prune the filters from convolutional layers. The network is trained for 300 epochs with the initial learning rate of 0.1 divided by 10 at epoch numbers 80, 140, and 230 to achieve the baseline accuracy. Similar to LeNet-5, a lower pruning ratio is used for initial layers compared to final layers. Consequently, 2\% of filters from layers 1 and 2 \textcolor{myblue}{(layers with 64 filters initially)}, 4\% of filters from layers 3 and 4 \textcolor{myblue}{(layers with 128 filters initially)}, 5\% of filters from layers 5, 6 and 7 \textcolor{myblue}{(layers with 256 filters initially)}, and 10\% of filters from the rest of the layers \textcolor{myblue}{(layers with 512 filters initially)} are pruned in each pruning iteration. After pruning, the network is retrained for 90 epochs beginning with a learning rate of 0.01, which is divided by 10 at epochs 40, and 70.
The pruning results for HRel-1 and HRel-2 (specified in Table \ref{tab:vgg_results}) are obtained for VGG-16 with 21-48-64-64-95-107-107-175-71-71-44-44-56 and 20-48-64-64-95-107-107-175-71-71-44-44-56 remaining filters, from each convolutional layer respectively. 

The HRel method achieves \textcolor{myblue}{the accuracy of 93.54\% when 84.70\% of the FLOPS pruned\%, which is better than all the other methods.} \textcolor{myblue}{In terms of accuracy drop HRel is observed to have second best result next to PP-OC. Also, a} very promising trade-off is observed between the accuracy and number of remaining FLOPs using the proposed HRel method as compared to the existing methods. It shows the capability of the proposed pruning method to prune a deeper plain model.

\begin{table}[t!]
\caption{Pruning results of LeNet-5 architecture over MNIST dataset. F here denotes number of remaining filters in convolutional layers 1 and 2 respectively.}
\label{tab:lenet_results}

\scriptsize
\bgroup
\def\arraystretch{1.1}%
\begin{tabu} to 0.5\textwidth {X[2.4]  X[1.6]  X[1.7]  X[0.9]  X[0.9]  X  X[0.9] }
\hline
\textbf{Method}&\textcolor{mymagenta}{\textbf{Acc$_{baseline}\%$}}&\textbf{Acc$_{pruned}$\%}& \textbf{\textcolor{mymagenta}{Acc$   \big\downarrow$}}  &\textbf{F} & \textbf{FLOPs}&\textbf{$P_f$\% \hspace{0.5cm} }\\ \hline
\hline
VIB \cite{dai2018compressing}&-&99.00&-& -& 0.09M& -\\
GAL \cite{lin2019towards}&99.20& 98.99&0.21&2, 15& 0.10M& 95.60\\
PP-OC \cite{singh2020acceleration} &99.17 & 99.20&-0.03&4, 5 & 0.19M & 95.56\\
\textbf{HRel-1(ours)}&\textbf{99.30}& \textbf{99.23}& \textbf{0.07} & \textbf{4, 5} & \textbf{0.19M} & \textbf{95.56} \\

\textbf{HRel-2(ours)} &\textbf{99.30}  &  \textbf{99.16}& \textbf{0.14}&\textbf{3, 5}  &\textbf {0.15M}& \textbf{96.41}\\

\textbf{HRel-3(ours)}         &\textbf{99.30} &  \textbf{98.99}& \textbf{0.31} & \textbf{3, 4} &\textbf{0.13M}& \textbf{96.84} \\

CFP \cite{singh2020leveraging}&99.17&  98.23 &0.94&2, 3 &0.08M&97.98 \\
HBFP \cite{basha2021deep}& 99.17& 98.60&0.57&2, 3 & 0.08M & 97.98 \\
\textbf{HRel-4(ours)}&\textbf{99.30}& \textbf{98.78} &\textbf{0.52}&\textbf{2, 3} & \textbf{0.08M} &\textbf{97.98}  \\
\hline
\end{tabu}
\egroup
\end{table}

\begin{table*}[!t]
\centering
\caption{Pruning results of VGG-16 architecture over CIFAR-10 dataset. Note that the entries are sorted based on the $P_f\%$ in increasing order.}
\scriptsize
\bgroup
\def\arraystretch{1.0}%
\begin{tabu} to \textwidth {X[2]  X[1.5]  X[1.5]  X  X[1.5]  X[1.5]  X[1.5]  X  }
\hline
\textbf{Method} &\textcolor{mymagenta}{\textbf{Acc$_{baseline}$\%}}&\textbf{Acc$_{pruned}$\%}& \textbf{\textcolor{mymagenta}{Acc$   \big\downarrow$}}   &  \textbf{FLOPs} & \textbf{$P_f$\%}  & \textbf{Params} & \textbf{$P_p$\%}  \\ \hline \hline
$\ell_1$-norm \cite{li2016pruning}&93.25 & 93.40 &\hspace{-1.5mm} -0.15 &206.00M &34.30& 5.40M&64.00  \\

Ayinde \textit{et al.} \cite{ayinde2019redundant}&93.80&  93.67 & 0.13  & - &40.50& -& 78.10\\

GAL \cite{lin2019towards}&93.96 &  90.78 &  3.18&171.89M &45.20& 2.67M &82.20 \\

CPGMI \cite{lee2020channel}&-&  93.86&-& 151.00M &51.80& 1.99M &86.70 \\
\textcolor{mygreen}{ABCPruner \cite{lin2020channel}} & \textcolor{mygreen}{93.02}&\textcolor{mygreen}{93.08}&\hspace{-1mm}\textcolor{mygreen}{-0.06} &\textcolor{mygreen}{82.81M} &\textcolor{mygreen}{73.68} &\textcolor{mygreen}{1.67M} &\textcolor{mygreen}{88.68}\\
CafeNet-E \cite{su2020locally}& - &93.67 &- & 76.00M & - & 1.40M& -\\ 
HRank \cite{lin2020hrank}&93.96& 91.23&2.73& 73.70M &76.50& 1.78M &92.00\\
VIB \cite{dai2018compressing}&-& 91.50&-& 70.63M & 77.48& -&-\\ 

MINT \cite{ganesh2020mint}&93.98& 93.43&0.55& -& -& -&83.43\\ 
CFP \cite{singh2020leveraging}&93.49&  92.90&0.59& 56.70M& 81.93&   & -\\ 
HBFP \cite{basha2021deep}&93.96& 91.99&1.97& 51.90M& 83.42& 2.40M& 83.77\\ 
PP-OC \cite{singh2020acceleration}&93.49&  93.43&0.06& 48.80M &84.50& 0.86M &94.30\\

\textbf{HRel-1(ours)}& \textbf{93.90} & \textbf{93.54}& \textbf{0.36} & \textbf{47.98M} &\textbf{84.70} & \textbf{0.75M}& \textbf{94.98}\\
\textbf{HRel-2(ours)}& \textbf{93.90} & \textbf{93.40}& \textbf{0.50} & \textbf{47.51M} &\textbf{84.85} & \textbf{0.75M}& \textbf{94.98}\\
Jordao \textit{et al.} \cite{jordao2020deep}&93.30&  91.80&1.50& -&90.66& -&-\\

\hline
\vspace{0.5em}
\end{tabu}
\egroup
     \label{tab:vgg_results}
\end{table*}

\begin{table*}[!t]
\centering
\caption{Pruning results of ResNet-56 architecture over CIFAR-10 dataset.}
\scriptsize
\bgroup
\def\arraystretch{1.0}%
\begin{tabu} to \textwidth {X[2]  X[1.5]  X[1.5]  X  X[1.5]  X[1.5]  X[1.5]  X}
\hline
\textbf{Method} &\textcolor{mymagenta}{\textbf{Acc$_{baseline}$\%}}&\textbf{Acc$_{pruned}$\%}& \textbf{\textcolor{mymagenta}{Acc$   \big\downarrow$}}   &  \textbf{FLOPs} & \textbf{$P_f$\%}  & \textbf{Params} & \textbf{$P_p$\%}  \\ \hline \hline
$\ell_1$-norm \cite{li2016pruning}&93.04& 93.06&\hspace{-1mm}-0.02& 90.90M& 27.60& 0.73M& 14.10\\ 
Ayinde \etal \cite{ayinde2019redundant} &93.39& 93.12&0.27& 90.70M& 27.90& 0.65M& 23.70\\ 
MINT \cite{ganesh2020mint}&92.55& 93.02&\hspace{-1mm}-0.47& -& -& -& 55.39\\ 
\textcolor{myorange}{SFP \cite{he2018soft}} & \textcolor{myorange}{93.59} & \textcolor{myorange}{92.26} & 
\textcolor{myorange}{1.33} & 
\textcolor{myorange}{59.40M} & 
\textcolor{myorange}{52.60}
& \textcolor{myorange}{-}& \textcolor{myorange}{-}\\

\textcolor{myorange}{ASFP \cite{he2019asymptotic}} & \textcolor{myorange}{93.59} & \textcolor{myorange}{92.44 }& \textcolor{myorange}{1.15 }& \textcolor{myorange}{59.40M} & \textcolor{myorange}{ 52.60} & \textcolor{myorange}{-} & \textcolor{myorange}{-}\\ 

TAS \cite{dong2019network} & - & 93.69 & 
0.77 & 
59.50M & 
52.70
& -& -\\
\textcolor{myorange}{LFPC \cite{he2020learning}} & \textcolor{myorange}{93.59} & \textcolor{myorange}{93.34} & \textcolor{myorange}{0.25} & \textcolor{myorange}{59.10M} & \textcolor{myorange}{52.90} & \textcolor{myorange}{-} & \textcolor{myorange}{-}\\ 
\textcolor{mygreen}{ABCPruner} \cite{lin2020channel} & \textcolor{mygreen}{93.26} & \textcolor{mygreen}{93.23} & \textcolor{mygreen}{0.03} & \textcolor{mygreen}{58.54M} & \textcolor{mygreen}{54.13} & \textcolor{mygreen}{0.39M} & \textcolor{mygreen}{54.20}\\ 
Jordao \etal \cite{jordao2020deep}&-& 93.71&-& -& 57.06& -& -\\ 
GAL\cite{lin2019towards} &93.26& 90.36&2.90& 49.99M & 60.20 & 0.29M& 65.90\\ 

\textbf{HRel-1(ours)}&\textbf{93.80}& \textbf{93.19}& \textbf{0.61} & \textbf{47.57M}& \textbf{62.06}& \textbf{0.30M}& \textbf{63.76}\\ 
PP-OC \cite{singh2020acceleration}&93.10& 93.15&\hspace{-1mm}-0.05& -& 68.40& -& -\\ 
Hrank \cite{lin2020hrank}&93.26& 90.72&2.54& 32.52M& 74.10& 0.27M& 68.10\\ 
CFP \cite{singh2020leveraging}&93.57& 92.63&0.93& 29.50M& 76.59& -& -\\ 
HBFP \cite{basha2021deep}&93.26&91.42&1.84& 27.10M& 78.43& 0.19M& 76.97\\ 

\textbf{HRel-2(ours)}&\textbf{93.80}& \textbf{92.70}& \textbf{1.10}& \textbf{28.99M}& \textbf{76.89}& \textbf{0.18M}& \textbf{77.83}\\

\hline
\vspace{0.5em}
\end{tabu}
\egroup
    \label{tab:res56_results}
\end{table*}

\begin{table*}[!t]
\centering
\caption{Pruning results of ResNet-110 architecture over CIFAR-10 dataset.}
\scriptsize
\bgroup
\def\arraystretch{1.0}%
\begin{tabu} to \textwidth {X[2]  X[1.5]  X[1.5]  X  X[1.5]  X[1.5]  X[1.5]  X}
\hline
\textbf{Method} &\textcolor{mymagenta}{\textbf{Acc$_{baseline}$\%}}&\textbf{Acc$_{pruned}$\%}& \textbf{\textcolor{mymagenta}{Acc$   \big\downarrow$}}   &  \textbf{FLOPs} & \textbf{$P_f$\%}  & \textbf{Params} & \textbf{$P_p$\%} \\ \hline
\hline
$\ell_1$ - norm \cite{li2016pruning}&93.53&93.30&0.23& 155.00M& 38.70& 1.16M& 32.60\\ 
Ayinde \etal \cite{ayinde2019redundant} &93.65&93.27&0.38 & 154.00M& 39.10& 1.13M& 34.20\\ 
\textcolor{myorange}{SFP \cite{he2019asymptotic}} & \textcolor{myorange}{93.68} & \textcolor{myorange}{93.38} & \textcolor{myorange}{0.30} & \textcolor{myorange}{150.00M} & \textcolor{myorange}{40.80} & \textcolor{myorange}{-} & \textcolor{myorange}{-}\\ 

GAL \cite{lin2019towards}&93.35& 92.55&0.80 & 130.20M& 48.50& 0.95M& 44.80\\ 
\textcolor{myorange}{ASFP \cite{he2019asymptotic}} & \textcolor{myorange}{93.68} & \textcolor{myorange}{93.10 }& \textcolor{myorange}{0.58 }& \textcolor{myorange}{121.00M} & \textcolor{myorange}{ 52.30} & \textcolor{myorange}{-} & \textcolor{myorange}{-}\\
Jordao \etal \cite{jordao2020deep}&-& 93.75&\hspace{2mm}-& \hspace{2mm} - & 60.17& -& -\\

TAS \cite{dong2019network} & - & 94.33 & 0.64 & 119.00M & 53.00 & - & -\\
\textcolor{myorange}{LFPC \cite{he2020learning}} & \textcolor{myorange}{93.68} & \textcolor{myorange}{93.79} & \textcolor{myorange}{\hspace{-1.5mm} -0.11} & \textcolor{myorange}{101.00M} & \textcolor{myorange}{60.30} & \textcolor{myorange}{-} & \textcolor{myorange}{-}\\ 
\textbf{HRel-1(ours)}&\textbf{93.50}&\textbf{93.03}& \textbf{0.47}& \textbf{095.72M}& \textbf{62.14}& \textbf{0.62M}& \textbf{63.80}\\ 
\textcolor{mygreen}{ABCPruner} \cite{lin2020channel} & \textcolor{mygreen}{93.50} & \textcolor{mygreen}{93.58} &\hspace{-1.5mm} \textcolor{mygreen}{-0.08} & \textcolor{mygreen}{089.87M} & \textcolor{mygreen}{65.04} & \textcolor{mygreen}{0.56M }& \textcolor{mygreen}{67.41}\\ 
HRank \cite{lin2020hrank}&93.50&92.65 &0.85& 079.30M& 68.60& 0.53M& 68.70\\ 
HBFP \cite{basha2021deep}&93.50&91.96 &1.54& 063.30M& 74.95& 0.43M& 74.92\\ 
\textbf{HRel-2(ours)}&\textbf{93.50}&\textbf{92.71}& \textbf{0.79} & \textbf{058.20M}& \textbf{76.95}& \textbf{0.38M}& \textbf{77.86}\\ 
\hline
\vspace{0.5em}
\end{tabu}
\egroup
    \label{tab:res110_results}
\end{table*}

\begin{table*}[!t]
\centering
\caption{Pruning results of ResNet-50 architecture over ImageNet dataset.}
\scriptsize
\bgroup
\def\arraystretch{1.0}%

\begin{tabu} to \textwidth {X[3] X  X[2]  X[1.5]  X[1.5]  X[2] X[1.5]  X X X X }

\hline
\textbf{Method} &\textcolor{mymagenta}{\textbf{Top-1}} &\textbf{Top-1$_{pruned}$\%}   & \textbf{\textcolor{mymagenta}{Top-1$\big\downarrow$}} &\textcolor{mymagenta}{\textbf{Top-5}} &\textbf{Top-5$_{pruned}$\%}   & \textbf{\textcolor{mymagenta}{Top-5$\big\downarrow$}} &  \textbf{FLOPs} & \textbf{$P_f$\%}  & \textbf{Params} & \textbf{$P_p$\%} \\ \hline
\hline
\textcolor{myorange}{SFP \cite{he2018soft} }& \textcolor{myorange}{76.15} & \textcolor{myorange}{62.14} & \textcolor{myorange}{14.01} & \textcolor{myorange}{92.87} & \textcolor{myorange}{84.60} &  \textcolor{myorange}{08.27 }& \textcolor{myorange}{- }& \textcolor{myorange}{41.80} & \textcolor{myorange}{-} & \textcolor{myorange}{-}\\
\textcolor{myorange}{ASFP \cite{he2019asymptotic}} & \textcolor{myorange}{76.15 }& \textcolor{myorange}{75.53 }& \textcolor{myorange}{00.62 }& \textcolor{myorange}{92.87 }& \textcolor{myorange}{92.73 }& \textcolor{myorange}{00.14}& \textcolor{myorange}{-}& \textcolor{myorange}{41.80} & \textcolor{myorange}{-} & \textcolor{myorange}{-}\\
GAL \cite{lin2019towards}&76.15&71.95 & 04.20 &92.87 & 90.94&01.93 & 02.33B & 43.03&21.20M&16.86\\
HRank \cite{lin2020hrank}&76.15&74.98 & 01.17 &92.87 & 92.33&00.54 & 02.30B & 43.76&16.15M&36.80\\ 

TAS \cite{dong2019network}&-&76.20 & 01.26 &- & 93.07&0.48&02.31B&43.50&-&-\\

DMCP \cite{guo2020dmcp}& 76.60 & 76.20 & 00.40 & - & - & - & 02.20B & 46.47 & - & -\\

\textbf{HRel-1(ours)} & \textbf{76.15} & \textbf{75.47} & \textbf{00.68} & \textbf{92.87} & \textbf{92.60} &  \textbf{00.27} & \textbf{02.11B} & \textbf{48.66} & \textbf{13.23M} & \textbf{48.24}\\




CafeNet-E \cite{su2020locally}& 77.80 & 76.90 & 00.90 & - & 93.10 & - & 02.00B & 51.33 & 18.40M & 27.84\\

\textcolor{mygreen}{MetaPruning} \cite{liu2019metapruning}&\textcolor{mygreen}{76.60}&\textcolor{mygreen}{75.40} & \textcolor{mygreen}{01.20} &\textcolor{mygreen}{- }& \textcolor{mygreen}{-}&\textcolor{mygreen}{-}&\textcolor{mygreen}{02.00B} & \textcolor{mygreen}{51.33 }&\textcolor{mygreen}{ -}&\textcolor{mygreen}{-}\\ 

PP-OC \cite{singh2020acceleration}&-&- & - &92.20 & 92.10&  00.10&- & -&15.70M&44.10\\ 
CFP \cite{singh2020leveraging}&-&- & - &92.20 & 91.40&  00.80&- & -&-&49.60\\


\textcolor{mygreen}{ABCPruner} \cite{lin2020channel} & \textcolor{mygreen}{76.01} & \textcolor{mygreen}{73.52} & \textcolor{mygreen}{02.49} & \textcolor{mygreen}{92.96 }& \textcolor{mygreen}{91.51} & \textcolor{mygreen}{01.45} & \textcolor{mygreen}{01.79B} & \textcolor{mygreen}{56.61 }& \textcolor{mygreen}{11.24M} & \textcolor{mygreen}{56.01}\\ 
\textbf{HRel-2(ours)} & \textbf{76.15} & \textbf{74.54 } & \textbf{01.61} &\textbf{92.87} & \textbf{92.12} & \textbf{00.75} & \textbf{01.69B} & \textbf{58.88} & \textbf{10.82M} & \textbf{57.67}\\
\textcolor{myorange}{LFPC \cite{he2020learning}} & \textcolor{myorange}{76.15} & \textcolor{myorange}{74.46} & \textcolor{myorange}{01.69} & \textcolor{myorange}{92.87} & \textcolor{myorange}{92.04} & \textcolor{myorange}{00.83} & \textcolor{myorange}{-} & \textcolor{myorange}{60.80}& \textcolor{myorange}{-}& \textcolor{myorange}{-}\\

\textbf{HRel-3(ours)} & \textbf{76.15} & \textbf{73.67} & \textbf{02.48 } &\textbf{92.87} & \textbf{91.70} & \textbf{01.17} & \textbf{01.38B} & \textbf{66.42} & \textbf{09.10M} & \textbf{64.40}\\


\hline
\vspace{0.5em}
\end{tabu}
\egroup
    \label{tab:res50_results}
\end{table*}

\color{black}

\subsection{ResNet-56 on CIFAR-10 Dataset}
\label{results:resnet56}
ResNet-56 is a deeper and complex architecture compared to VGG-16. ResNet-56 has 55 convolutional layers and 1 Fully connected layer in total. Except for the first one, all convolutional layers are grouped into three different blocks, with each block having 18 convolutional layers. The number of filters in $1^{st}$, $2^{nd}$ and $3^{rd}$ blocks is 16, 32 and 64, respectively. The network is trained for 180 epochs with the initial learning rate of 0.1, which is divided by 10 at epoch numbers 91 and 136 to achieve the baseline accuracy. For pruning ResNet-56, we follow the same approach as in \cite{singh2020leveraging}, i.e., pruning 1, 2 and 4 filters from every convolutional layer belonging to $1^{st}$, $2^{nd}$ and $3^{rd}$ blocks, respectively. After pruning, the network is retrained for 100 epochs beginning with a learning rate of 0.01, which is divided by 10 at epochs 20 and 70. \textcolor{myblue}{The remaining number of filters in convolutional layers of each block are 10, 20, 38 for HRel-1 and 8, 15, 30 for HRel-2.} As shown in Table \ref{tab:res56_results}, after pruning 62.06\% of the FLOPs, the proposed HRel method achieves 93.19\% accuracy, higher than GAL \cite{lin2019towards} method. In HRel-2 the highest percentage of parameters \ie 77.83\% and FLOPs \ie 76.89\% are pruned, and the accuracy can be observed to be better than HRank, CFP, and HBFP methods. \textcolor{myblue}{PP-OC has the least accuracy drop. CFP and HRel methods have the next best accuracy drops with higher $P_f$\% than PP-OC. Also, HRel-2 has a lesser accuracy drop compared to Hrank and HBFP methods.}

\subsection{ResNet-110 on CIFAR-10 Dataset}
\label{results:resnet110}
ResNet-110 contains 109 convolutional layers and 1 Fully connected layer in total. Similar to ResNet-56, except for the first convolutional layer, all the remaining convolutional layers are grouped into three different blocks, but each block contains 36 convolutional layers, with 16, 32 and 64 filters, respectively. The network is trained for 240 epochs with the initial learning rate of 0.1, which is divided by 10 at epoch numbers 88, 160, and 190 to achieve the baseline accuracy. Similar to ResNet-56, 1, 2, and 4 filters are pruned from each convolutional layer of $1^{st}$, $2^{nd}$, and $3^{rd}$ blocks, respectively. Note that similar to other methods, the first convolutional layer is not pruned. After pruning, the network is retrained for 70 epochs beginning with a learning rate of 0.01, which is divided by 10 at epochs 30 and 50. \textcolor{myblue}{The remaining filters in the convolutional layer of each block for HRel-1 are 10, 20, and 38 and for HRel-2 are 8,15, and 30.} HRel-1 \textcolor{myblue}{reduces} 62.1\% of the FLOPs and achieves an accuracy of 93.03\%, while Jordao \etal \cite{jordao2020deep} \textcolor{myblue}{ and ABCPruner obtained better accuracies of 93.75\% and 93.79\% by pruning a slightly lesser percentage of filters \ie 60.17\% and 60.30 respectively. NAS based method TAS achieves the highest accuracy of 94.33\%. ABCPruner and LFPC have less accuracy drop compared to other methods with nearby $P_f$\% values.} In HRel-2, with 76.95\% pruned FLOPs and 77.86\% pruned parameters, higher accuracy and lower accuracy drop than HRank and HBFP methods are observed.

\begin{figure*}[ht!]
  \begin{subfigure}{0.43\textwidth}
  \includegraphics[width=\textwidth]{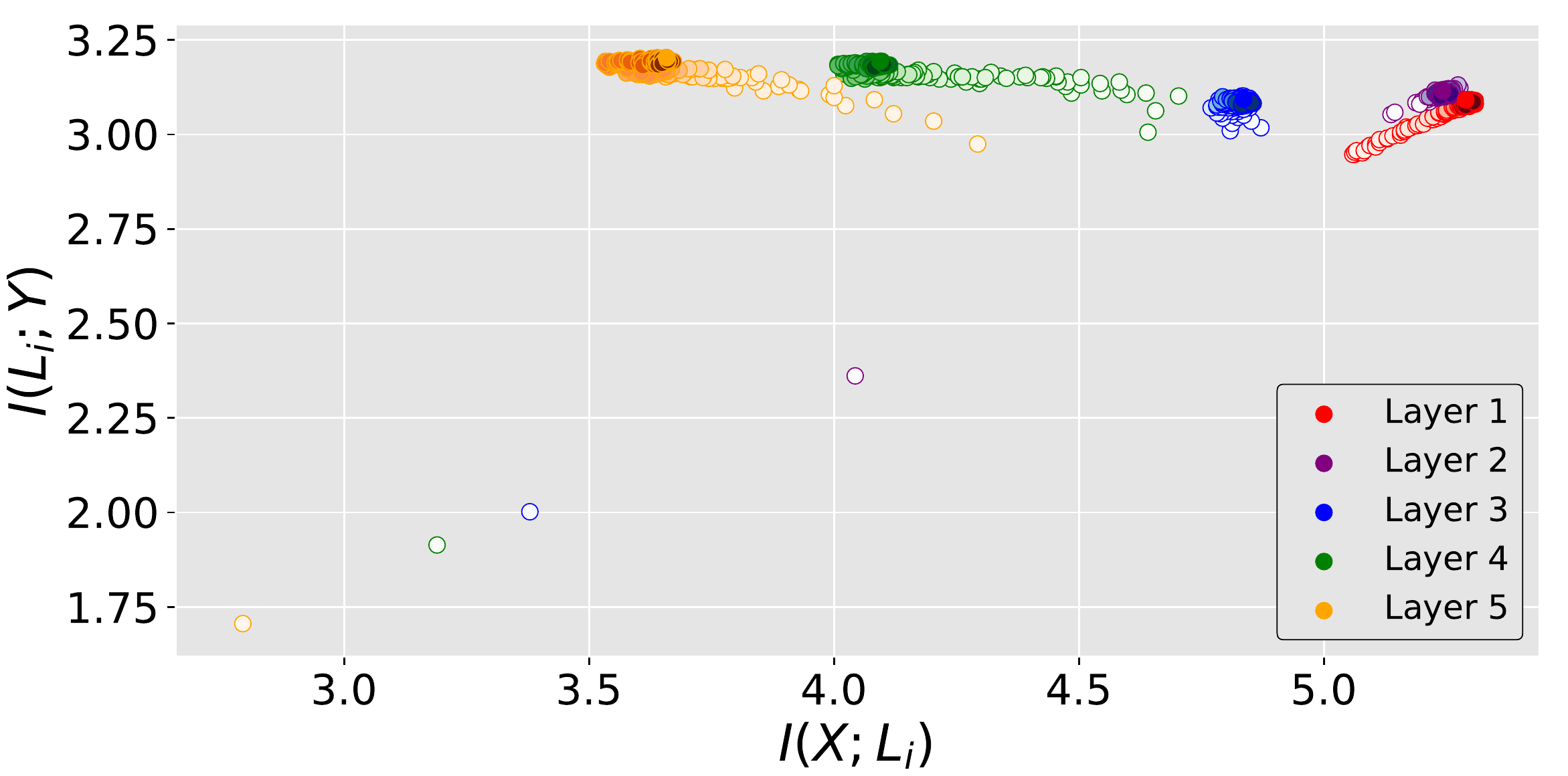}
  \caption{LeNet-5 Without pruning}
  \label{fig:lenet_ip_without_prune}
  \vspace*{0mm}
  \end{subfigure}
  \hfill
  \begin{subfigure}{0.43\textwidth}
  \includegraphics[width=\textwidth]{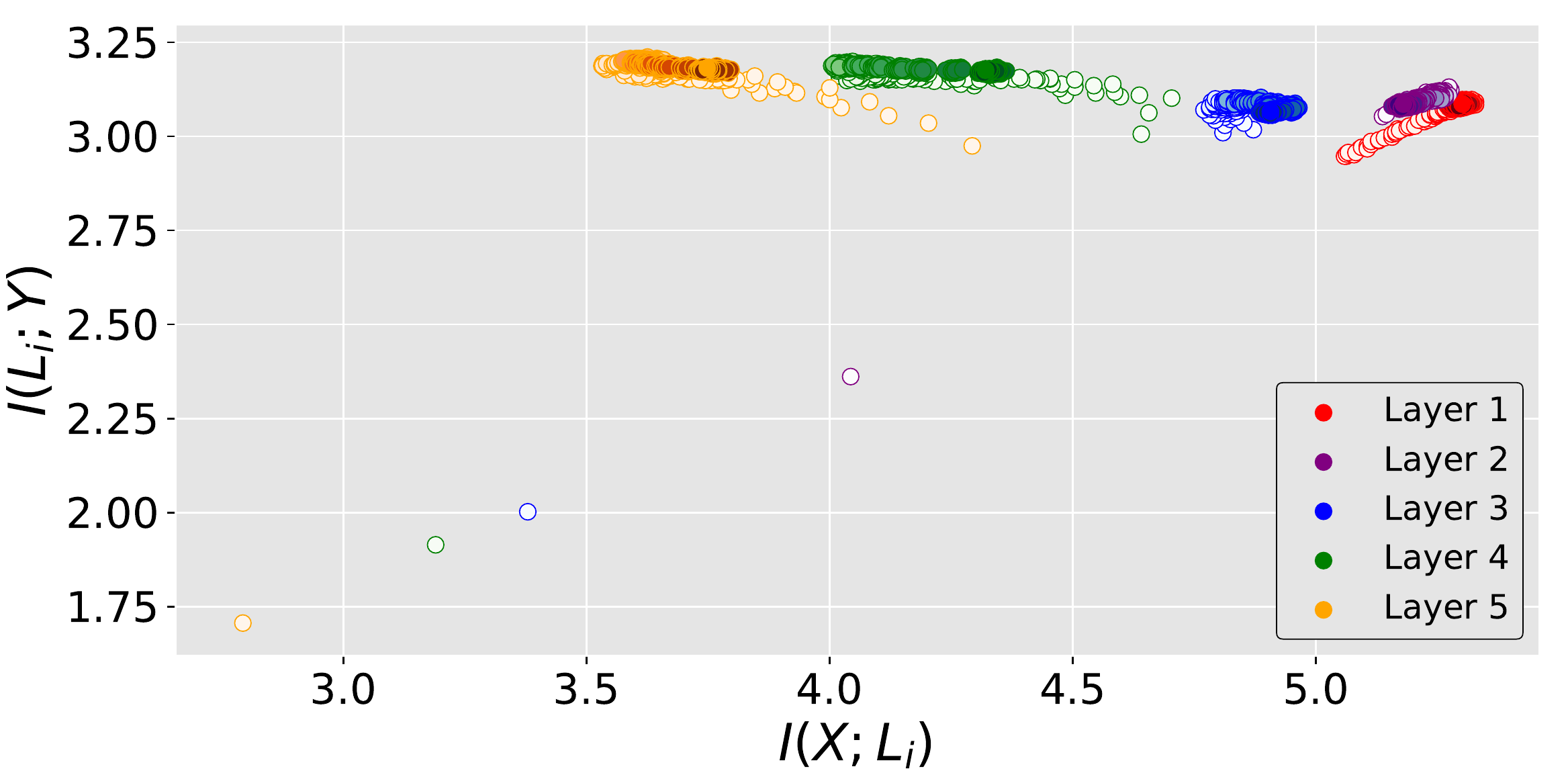}
  \caption{LeNet-5 With pruning}
  \label{fig:lenet_ip_with_prune}
  \vspace*{0mm}
  \end{subfigure} 
 \begin{subfigure}{0.43\textwidth}
  \includegraphics[width=\textwidth]{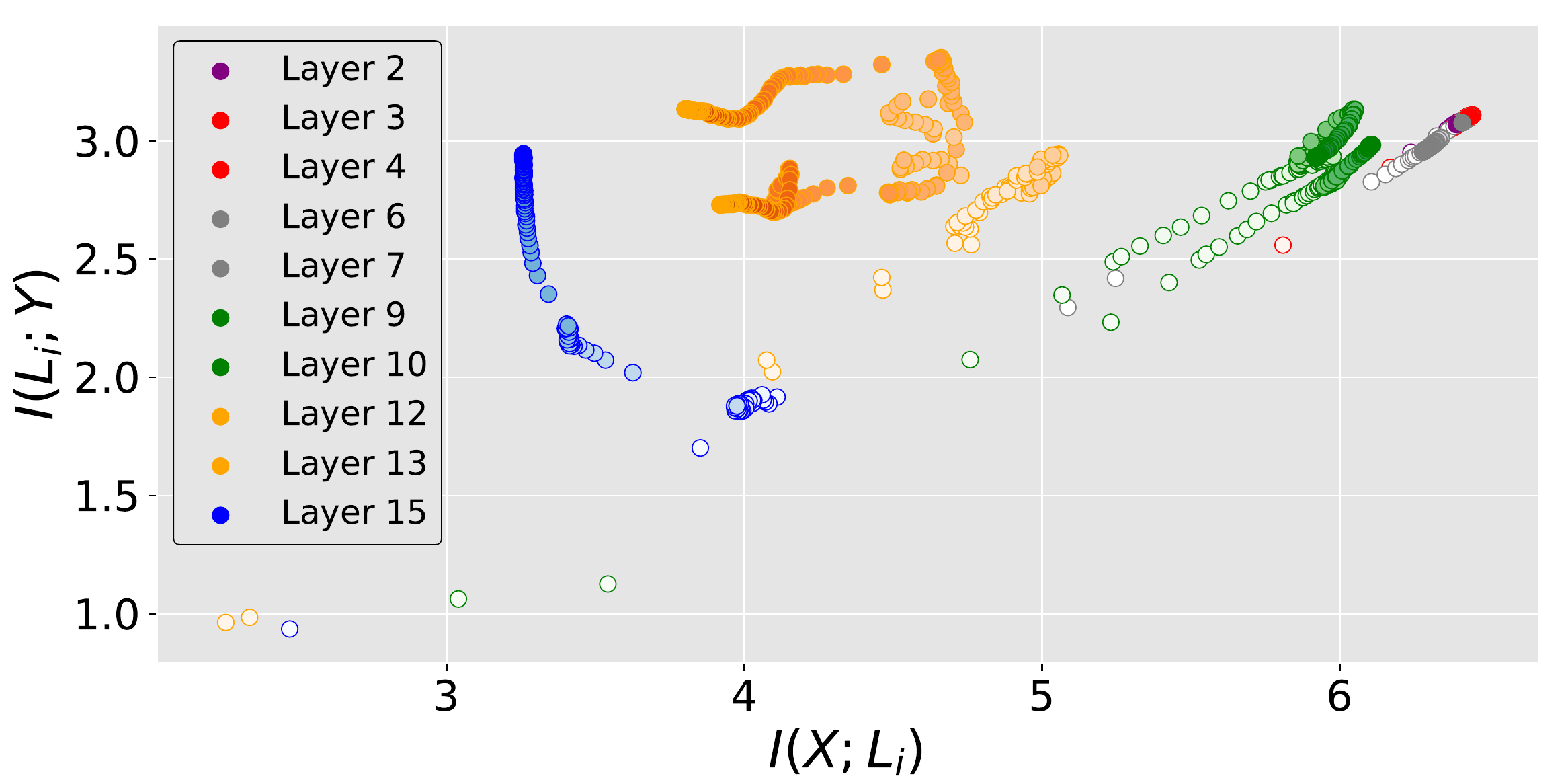}
  \caption{VGG-16 Without pruning}
  \label{fig:vgg_ip_without_prune}
  \vspace*{0mm}
  \end{subfigure}
  \hfill
  \begin{subfigure}{0.43\textwidth}
  \includegraphics[width=\textwidth]{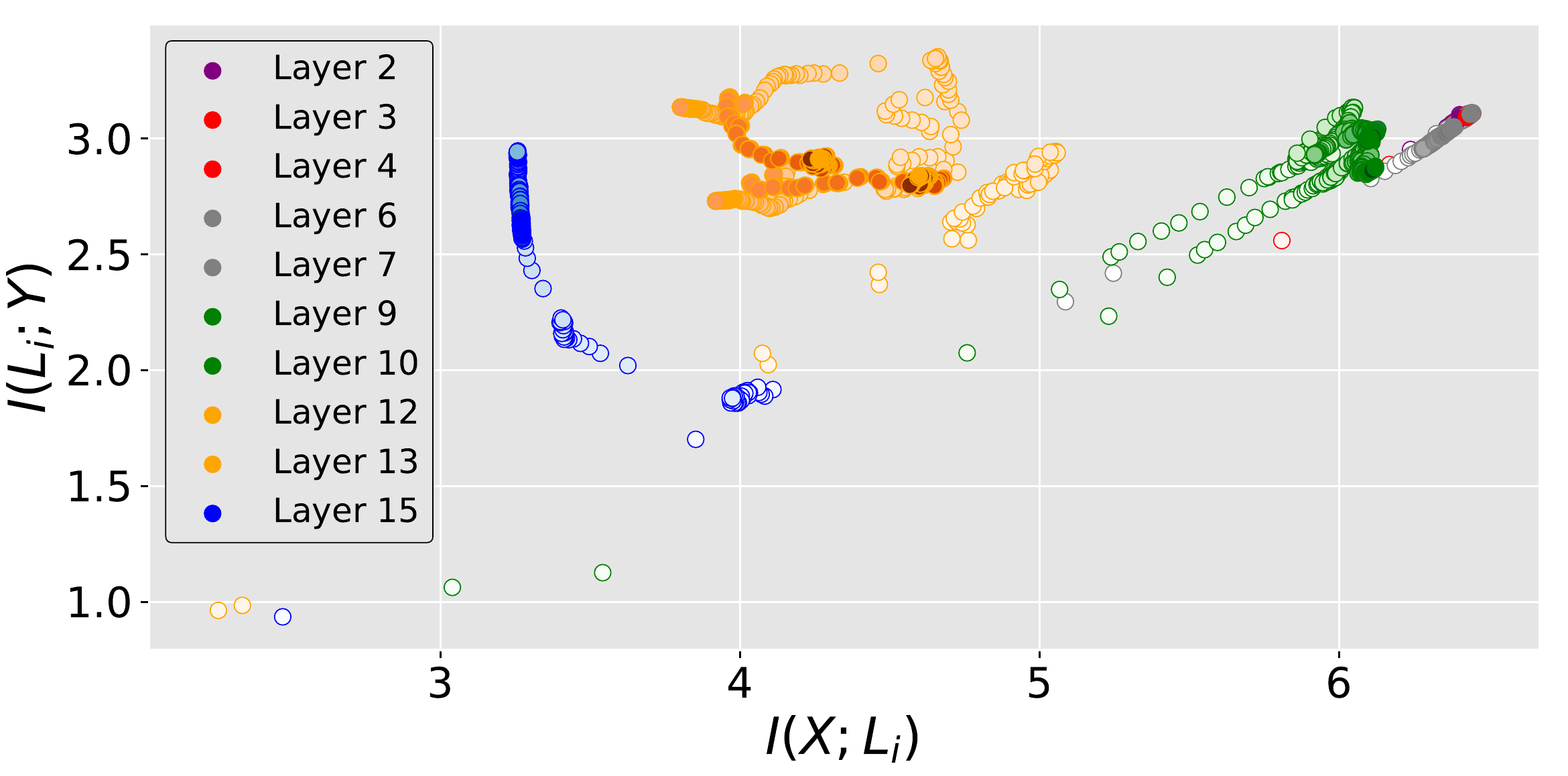}
  \caption{VGG-16 With pruning}
  \label{fig:vgg_ip_with_prune}
  \vspace*{0mm}
  \end{subfigure} 
  \begin{subfigure}{0.43\textwidth}
  \includegraphics[width=\textwidth]{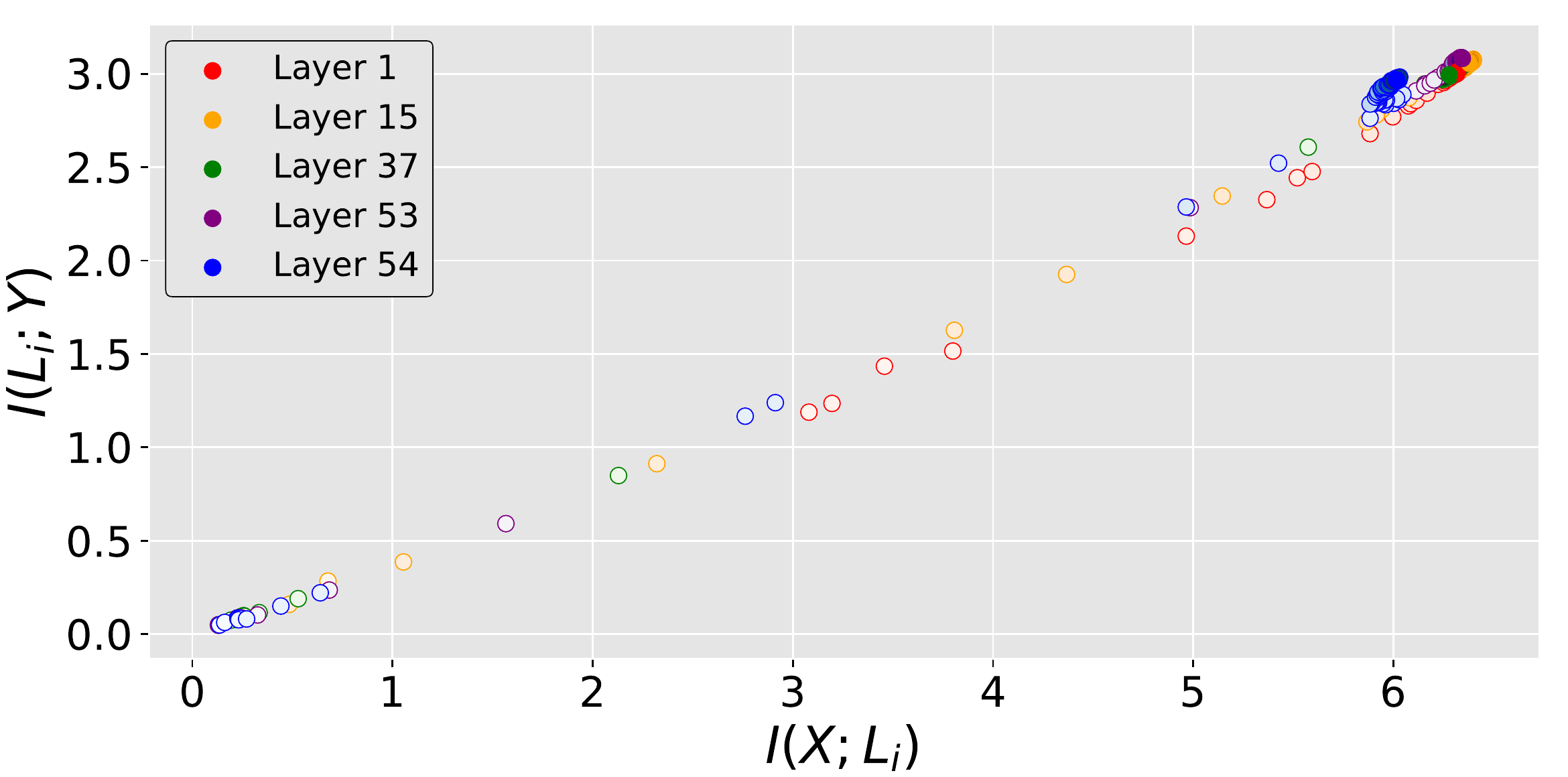}
  \caption{ResNet-56 Without pruning}
  \label{fig:resnet56_ip_without_prune}
  \vspace*{0mm}
  \end{subfigure}
  \hfill
  \begin{subfigure}{0.43\textwidth}
  \includegraphics[width=\textwidth]{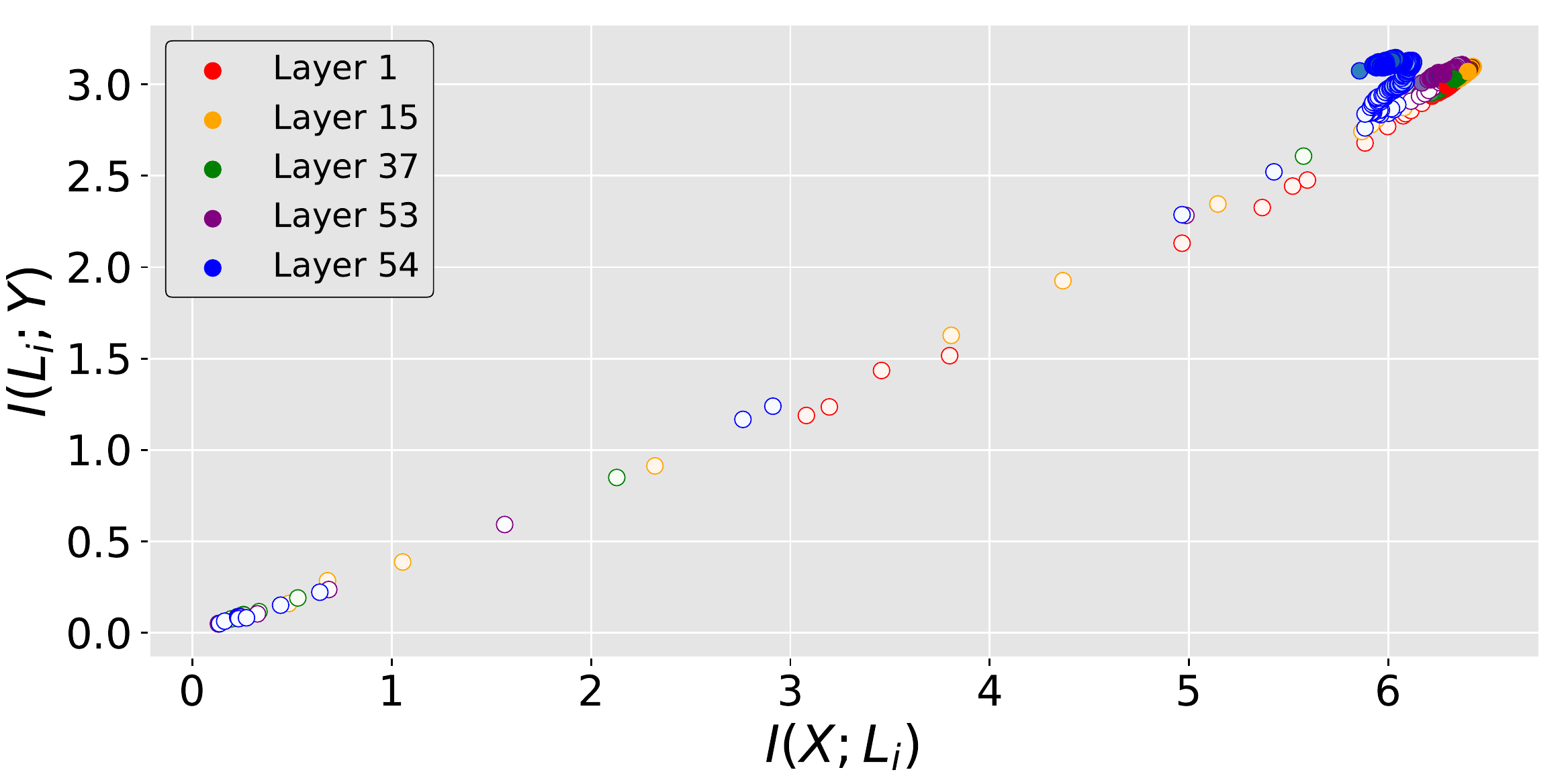}
  \caption{ResNet-56 With pruning}
  \label{fig:resnet56_ip_with_prune}
  \vspace*{0mm}
  \end{subfigure} 
  \begin{subfigure}{0.43\textwidth}
  \includegraphics[width=\textwidth]{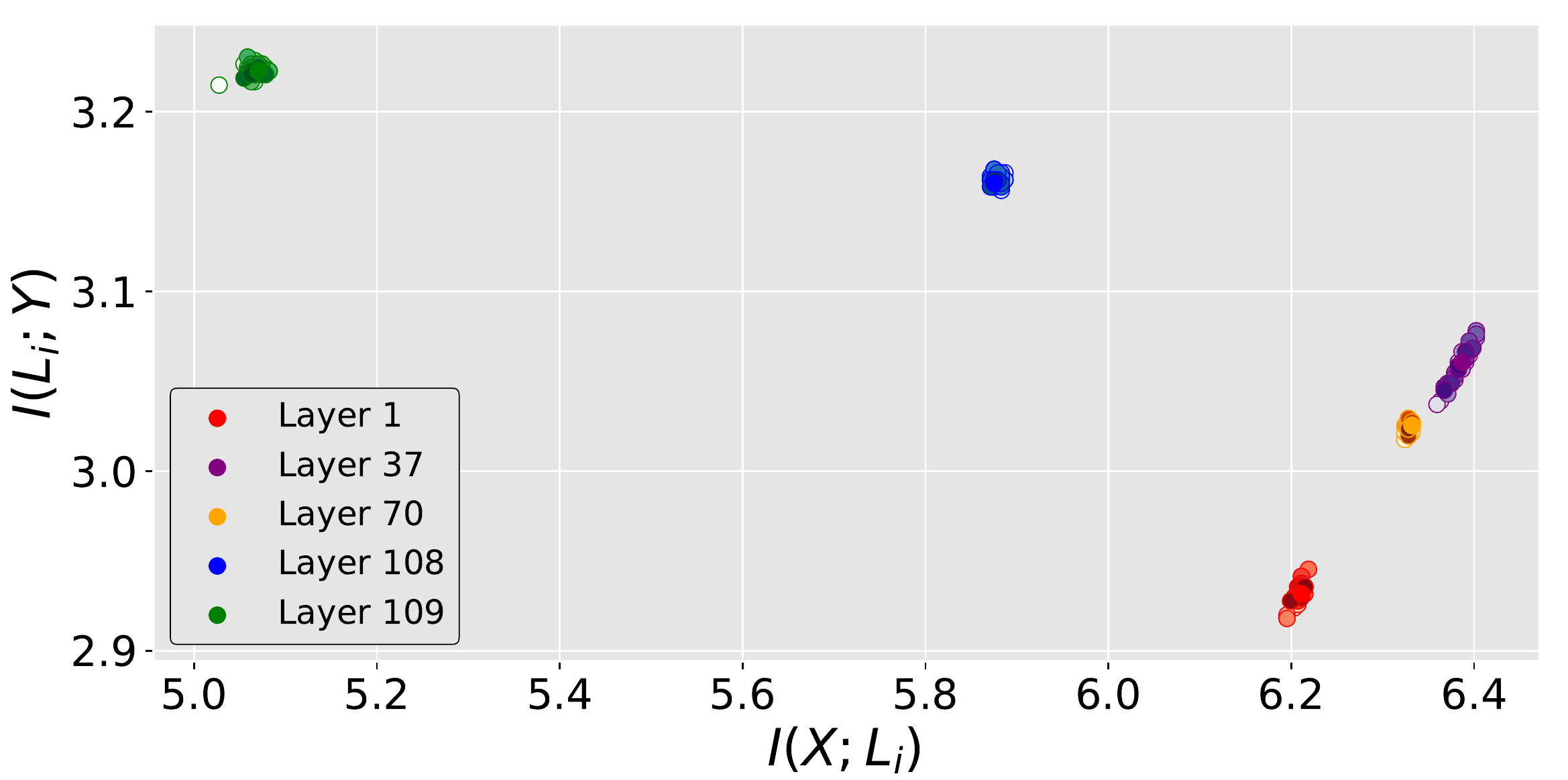}
  \caption{ResNet-110 Without pruning}
  \label{fig:resnet110_ip_without_prune}
  \vspace*{0mm}
  \end{subfigure}
  \hfill
  \begin{subfigure}{0.43\textwidth}
  \includegraphics[width=\textwidth]{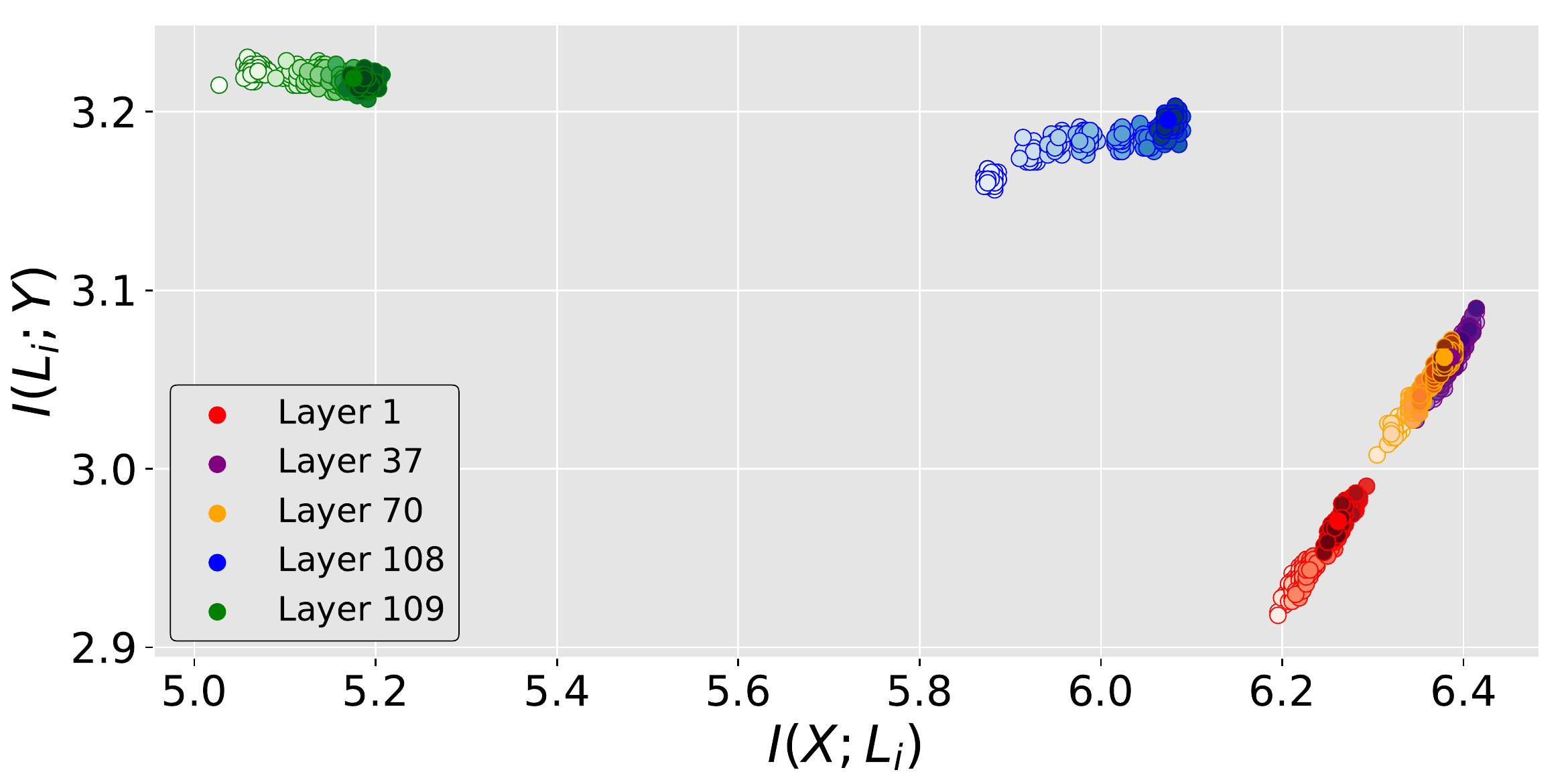}
  \caption{ResNet-110 With pruning}
  \label{fig:resnet110_ip_with_prune}
  \vspace*{0mm}
  \end{subfigure} 
  
  \begin{subfigure}{0.43\textwidth}
  \includegraphics[width=\textwidth]{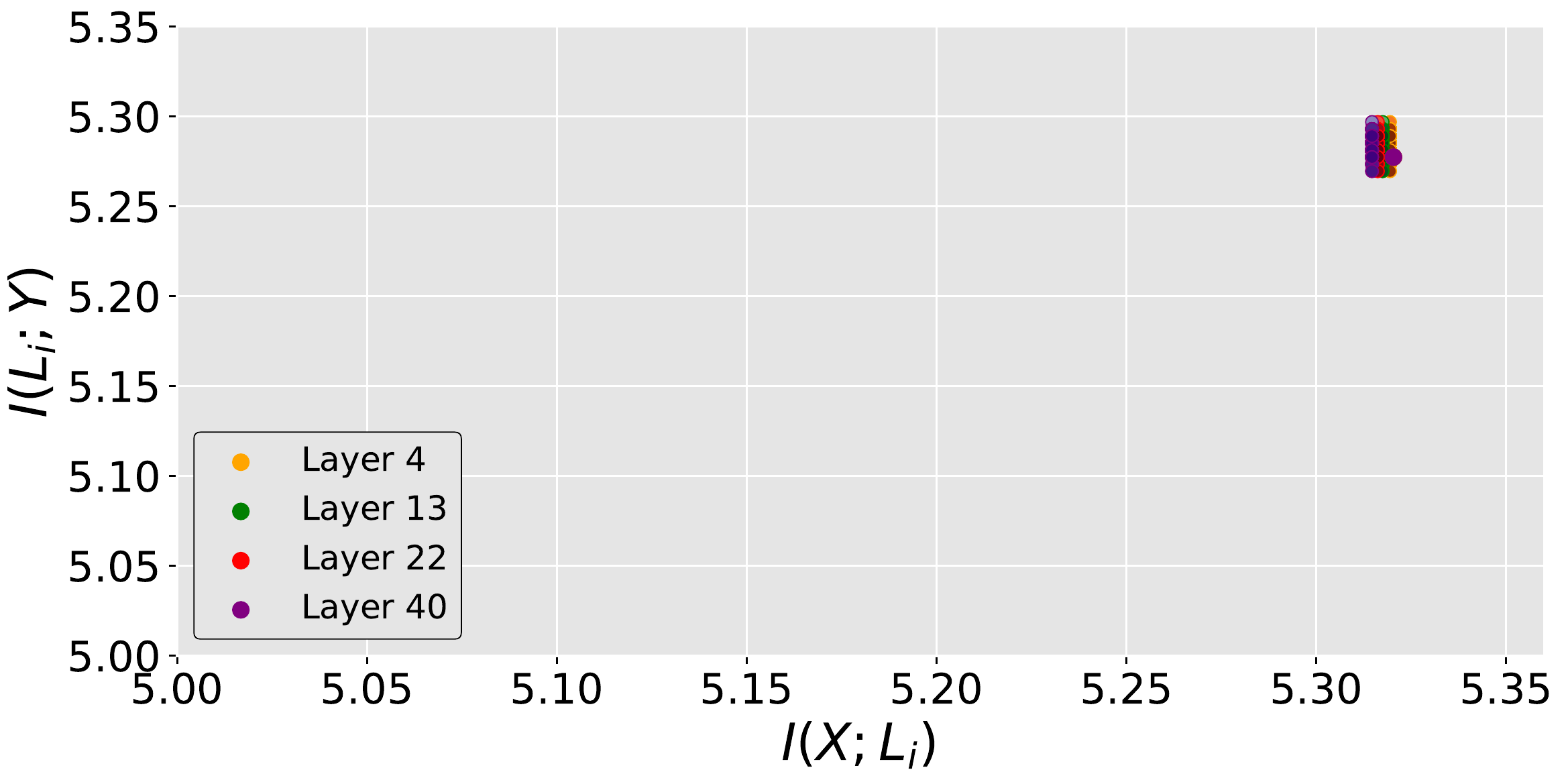}
  \caption{ResNet-50 Without pruning}
  \label{fig:resnet50_ip_without_prune}
  \vspace*{0mm}
  \end{subfigure}
  \hfill
  \begin{subfigure}{0.43\textwidth}
  \includegraphics[width=\textwidth]{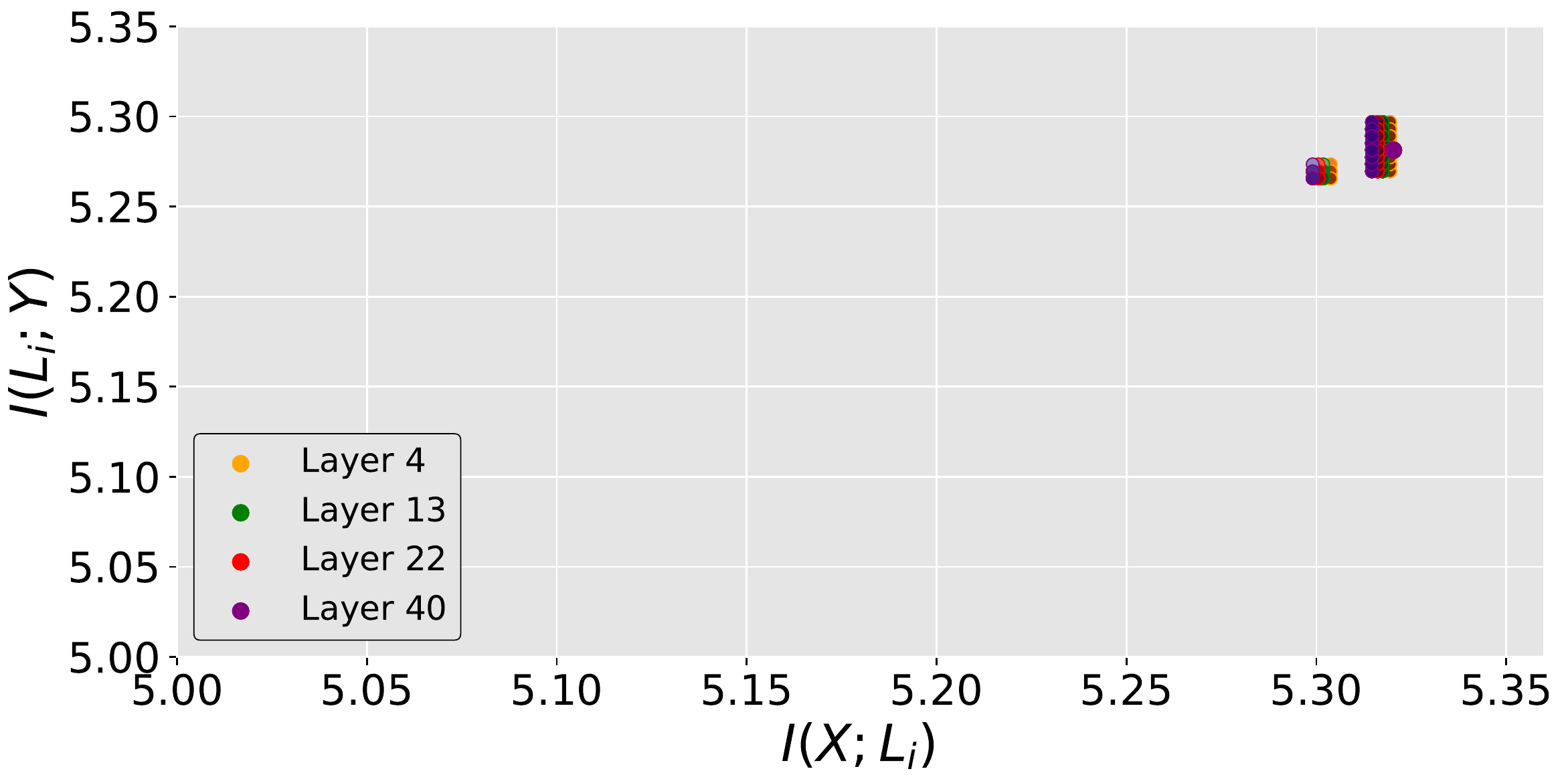}
  \caption{ResNet-50 With pruning}
  \label{fig:resnet50_ip_with_prune}
  \end{subfigure}

  \caption{Information Plane (IP) dynamics of LeNet-5, VGG-16, ResNet-56, ResNet-110, and ResNet-50 architectures. The left column corresponds to the IP dynamics of each architecture without pruning, and the right column shows the IP dynamics after pruning for the corresponding architecture. The layers are represented with different colors and saturation of each color indicates the progress of training.}
  \label{fig:ip_planes}
 \end{figure*}

 \begin{figure*}[!t]

     \centering
     \hspace{-3mm}
     \begin{subfigure}[b]{0.43\columnwidth}
         \centering
         \includegraphics[width=\textwidth,trim={0 2mm 0 0},clip]{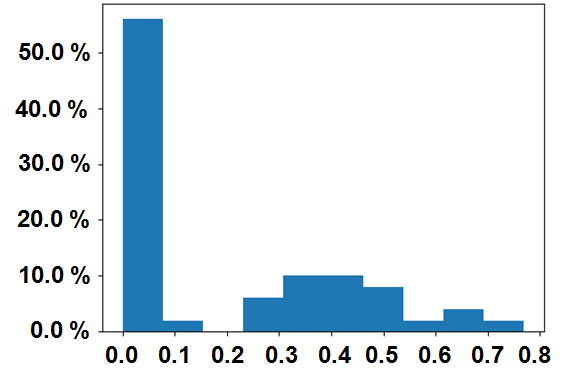}
         \caption{LeNet-5 with 50 Filters}
         \label{fig:dis_lenet_50}
     \end{subfigure}
     \hspace{-3mm}
     \begin{subfigure}[b]{0.43\columnwidth}
         \centering
         \includegraphics[width=\textwidth,trim={0 3mm 0 0},clip]{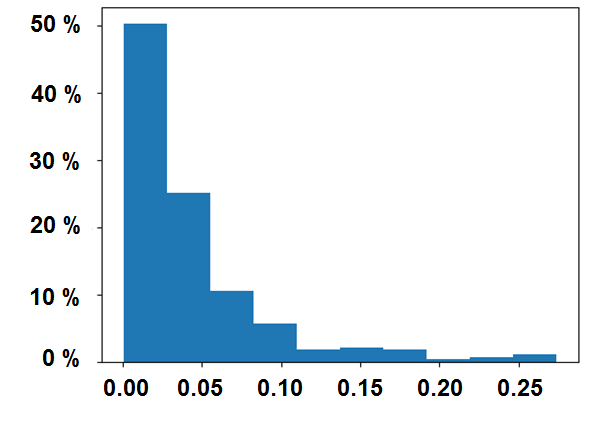}
         \caption{VGG-16 with 512 Filters}
         \label{fig:dis_vgg_512}
     \end{subfigure}
     \hspace{-3mm}
     \begin{subfigure}[b]{0.43\columnwidth}
         \centering
         \includegraphics[width=\textwidth]{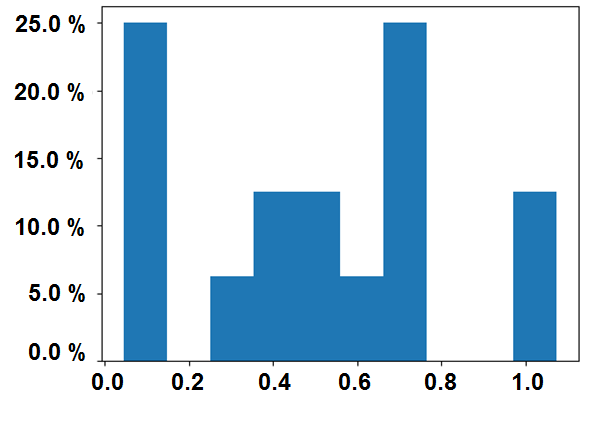}
         \caption{ResNet-56 with 16 Filters}
         \label{fig:dis_res56_16}
     \end{subfigure}
     \hspace{-3mm}
     \begin{subfigure}[b]{0.43\columnwidth}
         \centering
         \includegraphics[width=\textwidth,trim={0 2mm 0 0},clip,]{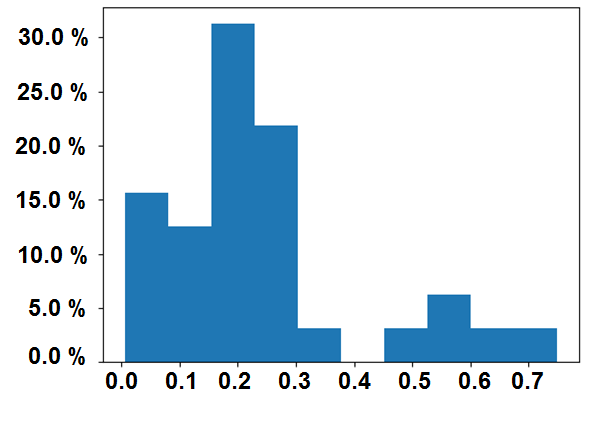}
         \caption{ResNet-110 with 32 Filters}
         \label{fig:dis_res110_32}
     \end{subfigure}
     \hspace{-3mm}
     \begin{subfigure}[b]{0.43\columnwidth}
         \centering
         \includegraphics[trim={0 0 0 1mm},clip,width=\textwidth]{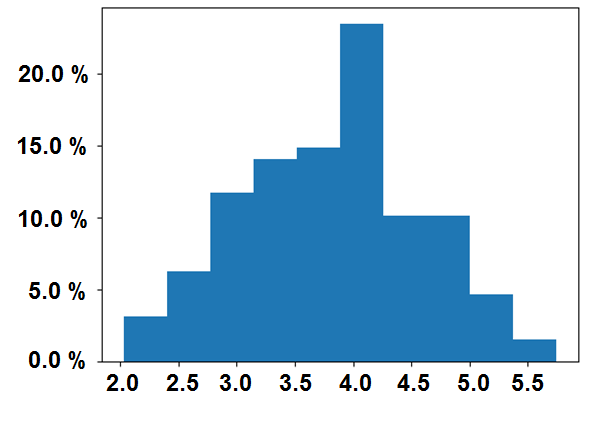}
         \caption{ResNet-50 with 128 Filters}
         \label{fig:dis_lenet_5}
         
     \end{subfigure}
     
     \hspace{-3mm}
     \begin{subfigure}[b]{0.43\columnwidth}
         \centering
         \includegraphics[width=\textwidth,trim={0 3mm 0 0},clip]{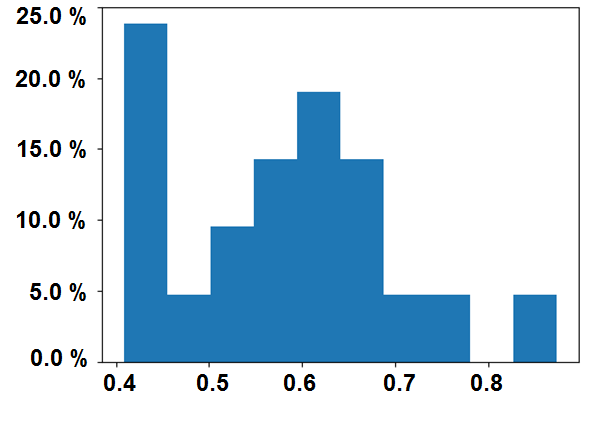}
         \caption{LeNet-5 with 21 Filters}
         \label{fig:dis_lenet_32}
     \end{subfigure}
     \hspace{-3mm}
     \begin{subfigure}[b]{0.43\columnwidth}
         \centering
         \includegraphics[width=\textwidth,trim={0 0mm 0 0},clip]{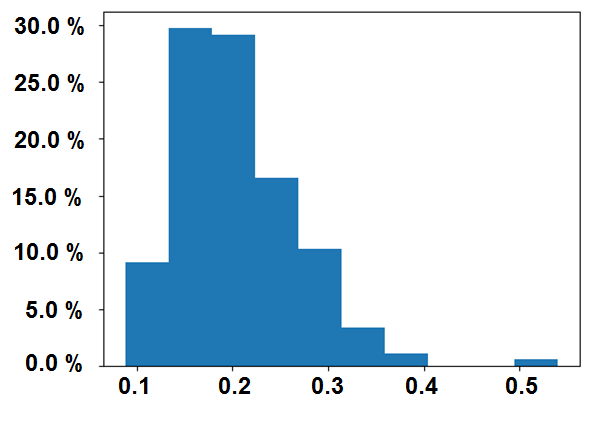}
         \caption{VGG-16 with 175 Filters}
         \label{fig:dis_vgg_175}
     \end{subfigure}
     \hspace{-3mm}
     \begin{subfigure}[b]{0.43\columnwidth}
         \centering
         \includegraphics[width=\textwidth]{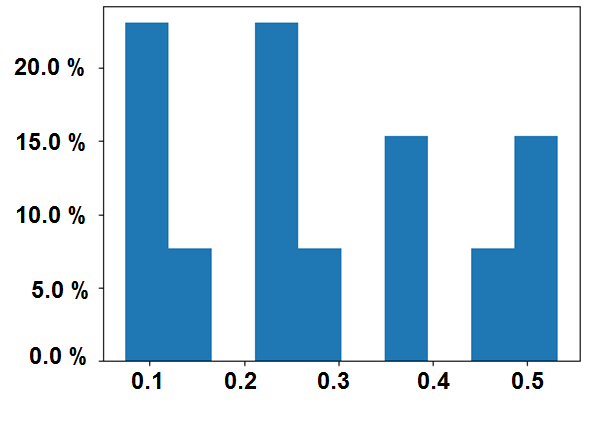}
         \caption{ResNet-56 with 13 Filters}
         \label{fig:dis_res56_13}
     \end{subfigure}
     \hspace{-3mm}
     \begin{subfigure}[b]{0.43\columnwidth}
         \centering
         \includegraphics[width=\textwidth]{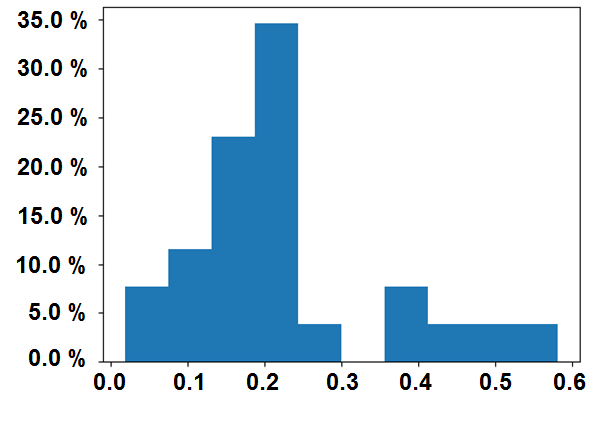}
         \caption{ResNet-110 with 26 Filters}
         \label{fig:dis_res110_71}
     \end{subfigure}
     \hspace{-3mm}
     \begin{subfigure}[b]{0.43\columnwidth}
         \centering
         \includegraphics[width=\textwidth]{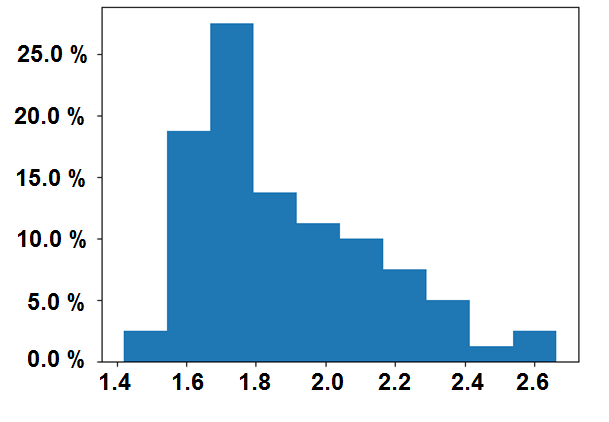}
         \caption{ResNet-50 with 80 Filters}
         \label{fig:dis_res50_80}
     \end{subfigure}

     \hspace{-3mm}
     \begin{subfigure}[b]{0.43\columnwidth}
         \centering
         \includegraphics[width=\textwidth]{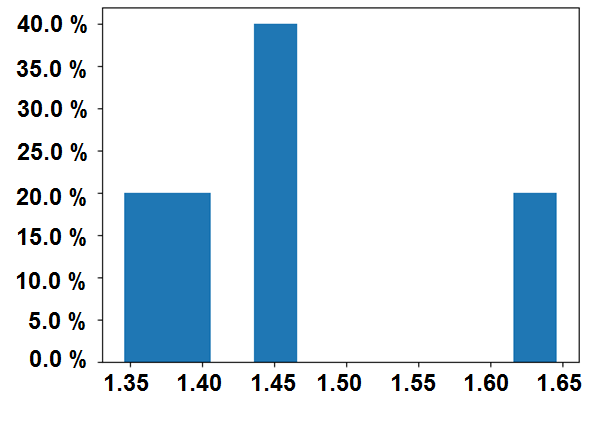}
         \caption{LeNet-5 with 5 Filters}
         \label{fig:dis_lenet_5}
     \end{subfigure}
     \hspace{-3mm}
     \begin{subfigure}[b]{0.43\columnwidth}
         \centering
         \includegraphics[width=\textwidth,trim={0 -2mm 0 0},clip]{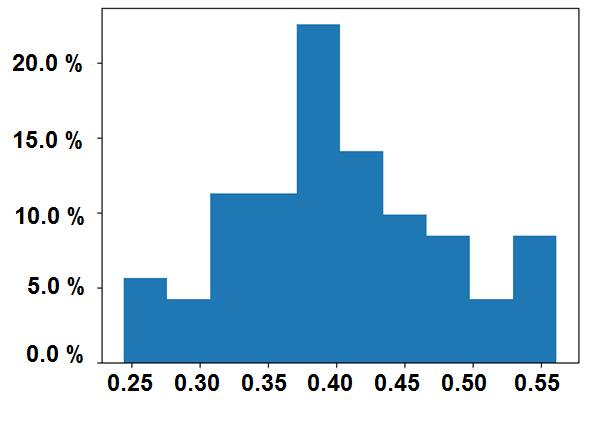}
         \caption{VGG-16 with 71 Filters}
         \label{fig:dis_vgg_71}
     \end{subfigure}
     \hspace{-3mm}
     \begin{subfigure}[b]{0.43\columnwidth}
         \centering
         \includegraphics[width=\textwidth]{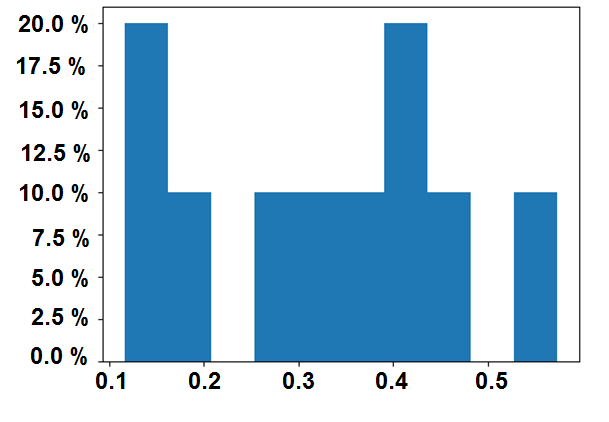}
         \caption{ResNet-56 with 10 Filters}
         \label{fig:dis_res56_10}
     \end{subfigure}
     \hspace{-3mm}
     \begin{subfigure}[b]{0.43\columnwidth}
         \centering
         \includegraphics[width=\textwidth]{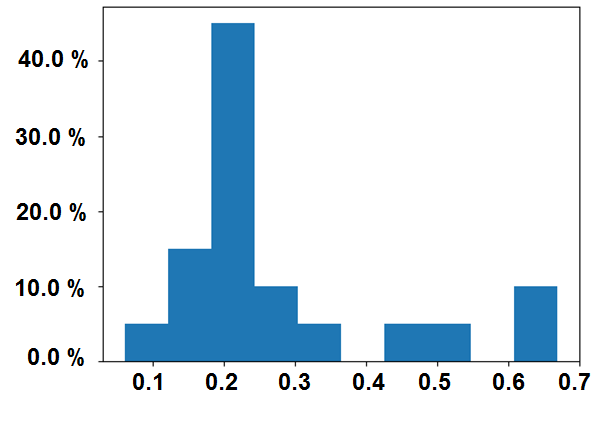}
         \caption{ResNet-110 with 20 Filters}
         \label{fig:dis_res110_20}
     \end{subfigure}
     \hspace{-3mm}
     \begin{subfigure}[b]{0.43\columnwidth}
         \centering
         \includegraphics[width=\textwidth]{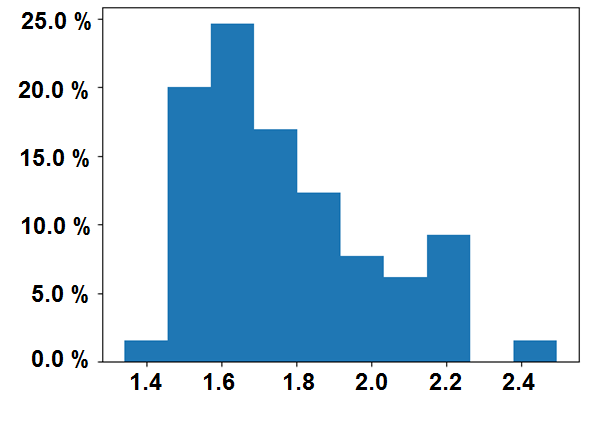}
         \caption{ResNet-50 with 65 Filters}
         \label{fig:dis_res50_65}
     \end{subfigure}

        \caption{Distribution of the Relevance values of remaining filters from convolutional layers of different architectures across the pruning iterations (the number of remaining filters is specified for each sub-figure). For each sub-figure, X-axis denotes the range of the Relevance values. Y-axis denotes the percentage of filters having a corresponding Relevance value range. Convolutional layers 2, 9, 15, 37, and 34 of LeNet-5, VGG-16, ResNet-56, ResNet-110, and ResNet-50 respectively are used.}
        \label{fig:Relevance_distribution}
\end{figure*}

\subsection{ResNet-50 on ImageNet}
\label{results:resnet50}
\textcolor{myblue}{ImageNet dataset consists of 1.2 million training images and 50,000 test images belonging to 1000 classes. ResNet-50 has 49 convolutional layers and 1 Fully connected layer in total. Except for the first one, all convolutional layers are grouped into four different blocks, with each block having two $1\times1$ convolutional layers and one regular convolutional layer (\ie $3\times3$ kernel size). The network is initialized with the pre-trained weights on the ImageNet dataset and trained for 3 epochs with a learning rate of 0.0001 (to learn the kernel width required for estimating MI at different layers). The 8\% of the convolutional layers from the first 3 blocks and 9\% from the last block are pruned in every pruning iteration. Similar to other approaches first two convolutional layers from every block are pruned. After pruning, the network is retrained for 33 epochs beginning with a learning rate of 0.001, divided by 10 at epochs 10 and 25. The remaining filters for HRel-1, HRel-2 and HRel-3 from the convolutional layers in each block are [41,80,158,288], [33,60,117,203] and [27,48,92,154], respectively. After pruning 58.88\% of the FLOPs, the proposed HRel method achieves 74.54\% Top-1 accuracy and 92.12\% Top-5 accuracy, with the least accuracy drop of $0.68$ among the network compression methods shown in Table \ref{tab:res50_results}. In HRel-3 the highest percentage of parameters \ie 64.40\% and FLOPs \ie 66.42\% are pruned and Top-1 accuracy of 73.67\% and Top-5 accuracy of 91.70\% are observed. HRel method shows comparable performance with MetaPruning and LFPC in terms of Top-1 and Top-5 accuracies. However, it achieves higher Top-1 and Top-5 accuracies than SFP, ASFP, GAL, HRank, PP-OC, CFP and ABCPruner for comparable $P_f$\%. In terms of accuracy drop, the HRel method has the lowest Top-1 accuracy drop and second-best Top-5 accuracy drop than other network compression methods with comparable $P_f$\%. Compared to all the methods, HRel has the second-best accuracy drop next to DMCP.}

\textcolor{myblue}{The results observed using ResNet-56, ResNet-110, and ResNet-50 on CIFAR-10 and ImageNet datasets} point out that the proposed HRel pruning method can prune the residual networks with very promising performance in terms of the accuracy as well as pruned FLOPs and parameters. Note that the proposed method has also shown very appealing performance for shallow (LeNet-5) and deep (VGG-16) plain models over different datasets. Overall, it can be deduced from the above experimental results that the proposed HRel pruning approach can retain the filters that are having high Relevance which leads to better accuracy even after pruning. The proposed pruning method is robust since high Relevance is used as the criterion in modeling the proposed HRel pruning strategy with the help of the IB theory.

\subsection{Analysis of Information Plane Dynamics}
\label{results:ip_plane}

IP dynamics for \textcolor{myblue}{the architectures} LeNet-5, VGG-16, and ResNet-56 are plotted using $I(X;L_i)$ and $I(L_i;Y)$ from the beginning of the training. Whereas for ResNet-110, values from the last ten epochs before reaching the baseline accuracy are observed due to its complex architecture. \textcolor{myblue}{For ResNet-50, the values are observed only from the last epoch during the initial training due to comparatively more mini-batches for the ImageNet dataset. After initial training, for every pruning iteration, the last one epoch for ResNet-50 and the last ten epochs for the rest of the architectures are used to estimate the values $I(X;L_i)$ and $I(L_i;Y)$.} From the IP dynamics of each architecture after pruning, \ie in the second column in Fig. \ref{fig:ip_planes}, a slight decrease in the Relevance value of the final layers is observed compared to the architecture's IP dynamics without pruning (which is very less in the case of ResNet-50), is observed. This means that the network layers lose slight information concerning the class labels. This can be related to the small accuracy drop resulted from the pruning of filters. Though we preserve the filters with high Relevance, based on ``Partial information decomposition" of MI \cite{Williams2010NonnegativeDO,yu2020understanding}, the unique information (i.e., the information provided individually by few pruned filters) and their synergy (i.e., joint information provided only by the combination of few filters) is lost.

\textcolor{myorange}{The MI estimator \cite{wickstrom2019information} used in the HRel method highly depends on the optimal kernel bandwidth of the dataset \cite{tapia2020information}. The original work \cite{wickstrom2019information} and a few related works \cite{yu2019multivariate,yu2020understanding} using this estimator have used only the smaller datasets like CIFAR-10, MNIST, Fruits 360 \cite{muresan2018fruit}, MADELON \cite{guyon2004result} etc. In these datasets, kernel bandwidth is chosen either by Silverman's rule of thumb \cite{silverman1986monographs} or by empirical evaluation over a range of values. Due to high dimensionality and more data samples in the ImageNet dataset, it is not feasible to obtain optimal kernel bandwidth for ImageNet using these methods. Hence, the same kernel bandwidth values specified in the MI estimator are used for input and class labels of the ImageNet dataset. Though similar and saturated values are observed for all layers in the Information Plane before and after pruning ResNet-50, the filters' Relevance values have shown enough variation among them as shown in Fig. \ref{fig:mi_over_epoch}, facilitating the selection of filters during pruning iterations. The optimal bandwidth for input and class labels of the ImageNet dataset can produce a better projection of the Information Plane of ResNet-50.}

Overall, the IP dynamics indicate some information loss while pruning the filters, which is minimal for the proposed HRel method, as indicated by the experimental results. Thus, minimal information loss is acceptable as the complexity of the model is reduced drastically to facilitate the deployment of deep learning models over resource-constrained devices.

\subsection{Analysis of Progression of Pruning using the Relevance Distribution}
\label{results:distribution}
The distribution of the Relevance values of each architecture at the beginning of certain pruning iterations is shown in Fig. \ref{fig:Relevance_distribution}. Each column represents the Relevance value distribution for all the remaining filters in a given layer of the architecture. The first row is the distribution of the Relevance value before the beginning of the first pruning iteration. Subsequent rows can be identified by the remaining filters mentioned for each architecture. Note that the plot in Fig. \ref{fig:Relevance_distribution} shows the percentage (\%) of remaining filters for different ranges of Relevance values. From the Fig. \ref{fig:dis_lenet_50} - \ref{fig:dis_res110_32}, \ref{fig:dis_lenet_32} - \ref{fig:dis_res110_20} and \ref{fig:dis_lenet_5} - \ref{fig:dis_res110_20} it is observed that in each column, with the increase in the pruning iterations, the lowest Relevance value among the remaining filters increased. Also, in few architectures such as ResNet-56 and ResNet-110, though the lowest Relevance value did not change much, the percentage of filters having the lowest Relevance value is comparatively decreased after pruning.

\begin{figure}[!t]
  \centering
  \includegraphics[scale=0.25]{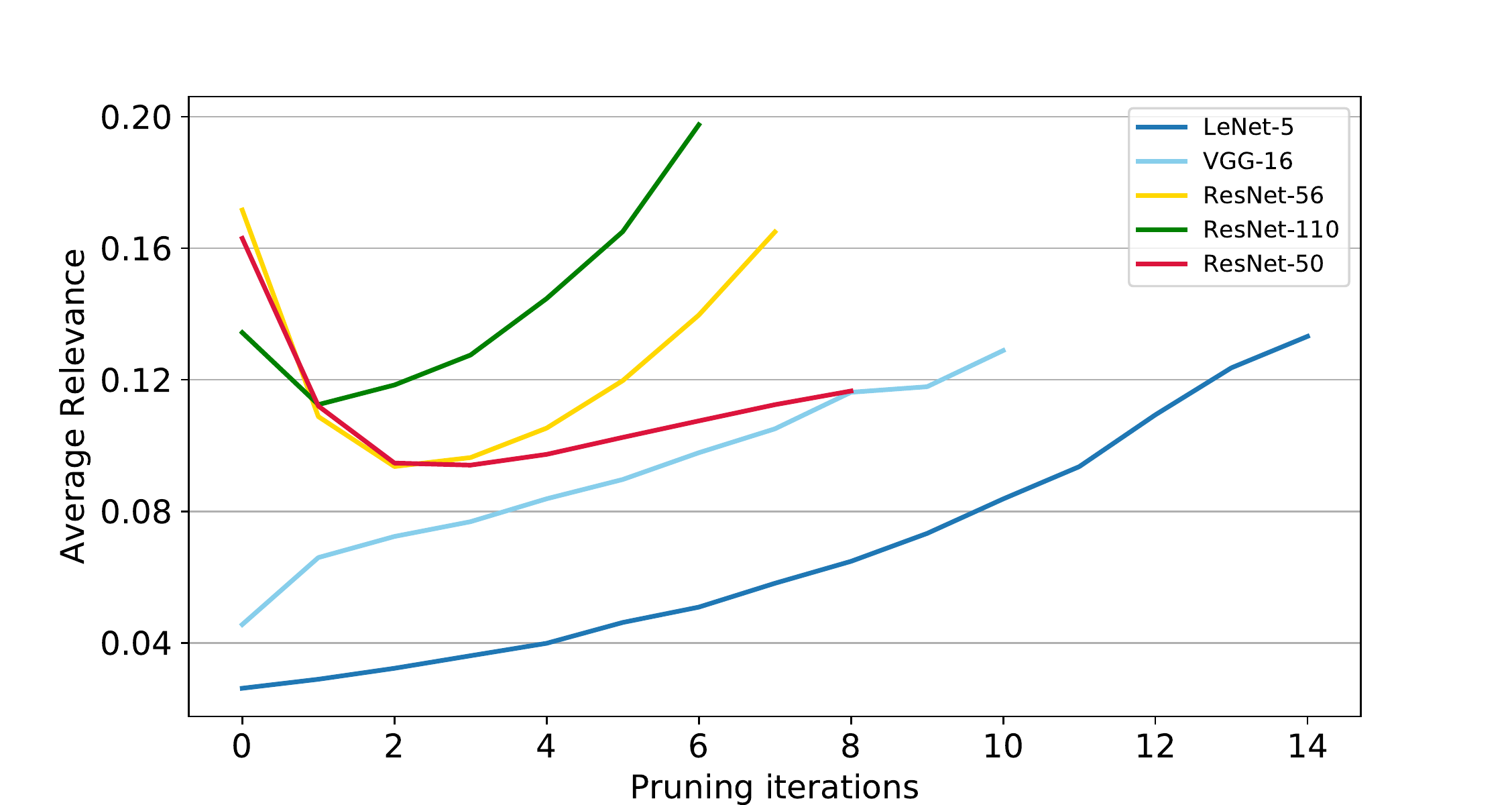} 
  \caption{The average Relevance of all the layers across the pruning iterations for various architectures.}
  \label{fig:average_relevance}
  \end{figure}

It can be noticed that the distribution is slightly shifted towards the right side in most of the cases across the pruning iterations, which shows that the Relevance of the majority of the remaining filters is high. From Fig. \ref{fig:average_relevance}. it can be observed that the average Relevance across the pruning iterations increased continuously for LeNet-5 and VGG-16. However, for ResNet-56 and ResNet-110 the average Relevance decreased initially for few pruning iterations 
and then increased. This observation further supports the proposed idea of utilization of high Relevance in the HRel pruning method. For ResNet-50 there is no much increment observed even after few pruning iterations in Fig. \ref{fig:average_relevance}. The Relevance distribution values are also shifted to the left for ResNet-50 in Fig. \ref{fig:Relevance_distribution}.  
The Relevance distribution values can also be more accurate if the optimal kernel bandwidth is used.

\subsection{Ablation study}
\label{results:ablation_study}
An ablation study is conducted to understand the effect of global filter pruning based on Relevance values and the effect of batch size during the estimation of filters' Relevance.

\begin{figure}[!t]
  \centering
  \includegraphics[scale=0.30]{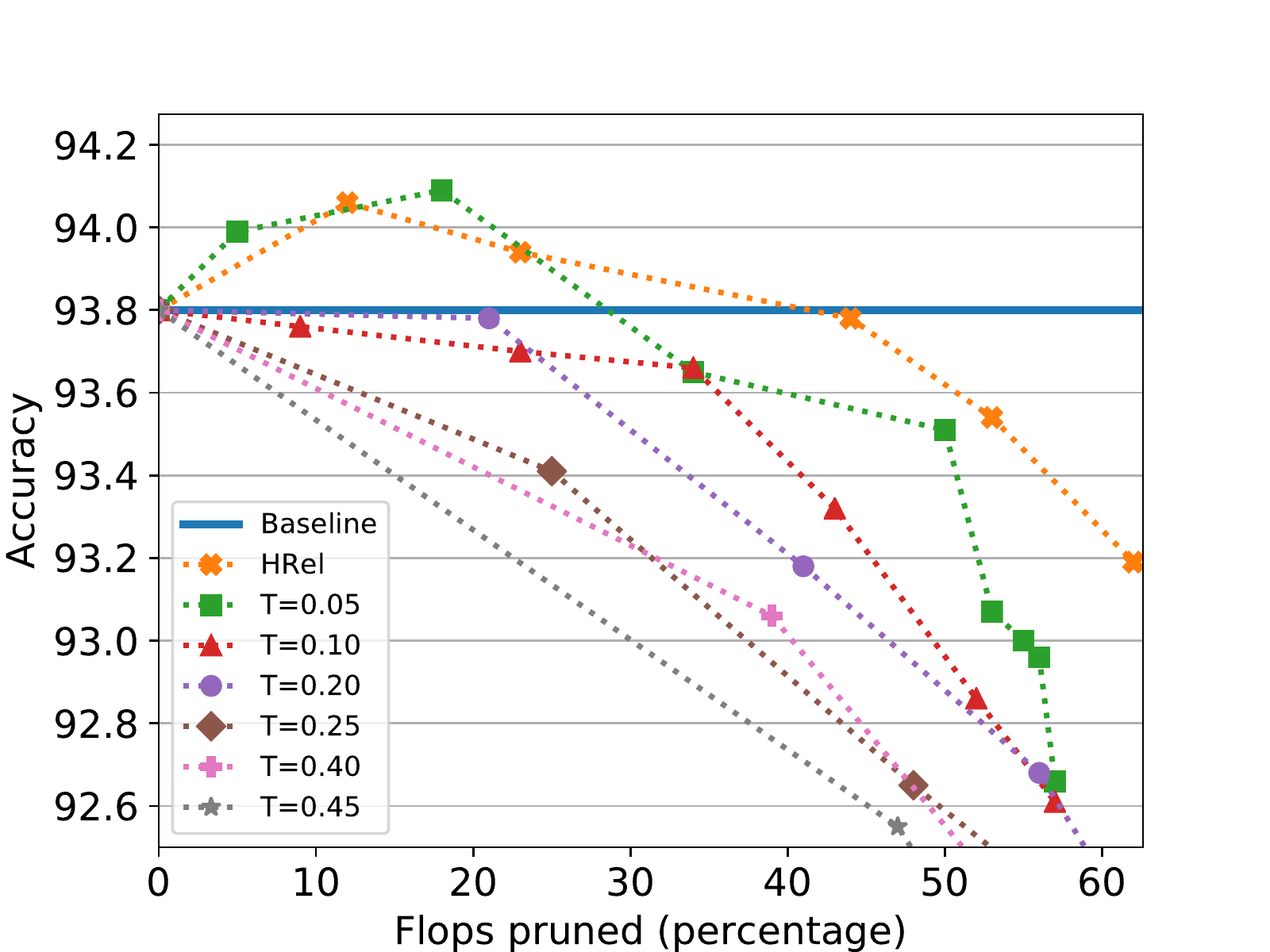} 
  \caption{Accuracy of ResNet-56 architecture on CIFAR10 dataset when filters are globally pruned.}
  \label{fig:global_pruning}
  \end{figure}

\textcolor{mygreen}{\textit{1) Global pruning}:
The filters are compared globally based on their Relevance values in the global pruning method. As shown in Fig. \ref{fig:Relevance_distribution}, filters' Relevance values keep changing with the pruning iterations. Thus, in every pruning iteration, the estimated Relevance values of all the remaining filters across the layers are sorted and the maximum value from the least T\% of the values is considered as the threshold. A new threshold value based on the filters' Relevance is used at every pruning iteration. Consequently, the filters with Relevance below the threshold are pruned. The experiments are conducted on ResNet-56 architecture using CIFAR-10 dataset with the values 5, 10, 20, 25, 40, and 45 for T. In the global pruning method, the lower T values resulted in better accuracy, as shown in Fig. \ref{fig:global_pruning}. However, it can be observed that none of the global pruning results have achieved comparable accuracy with the proposed HRel method, where filters are ranked layer-wise based on their Relevance. Even the recent work \cite{amjad2021understanding} concludes that it is not suggested to compare the filters' Relevance across layers. However, it is a good estimator for the filter's importance when compared layer-wise. The global pruning method in \cite{amjad2021understanding} prunes the filters at a higher rate from the layers with relatively low Relevance filters. A similar observation is found in our ablation study. Filters are selectively pruned from a few layers at a higher rate, due to which more accuracy drop is observed, as depicted in Fig. \ref{fig:global_pruning}.
}

\begin{figure}[!t]
  \centering
  \includegraphics[scale=0.30]{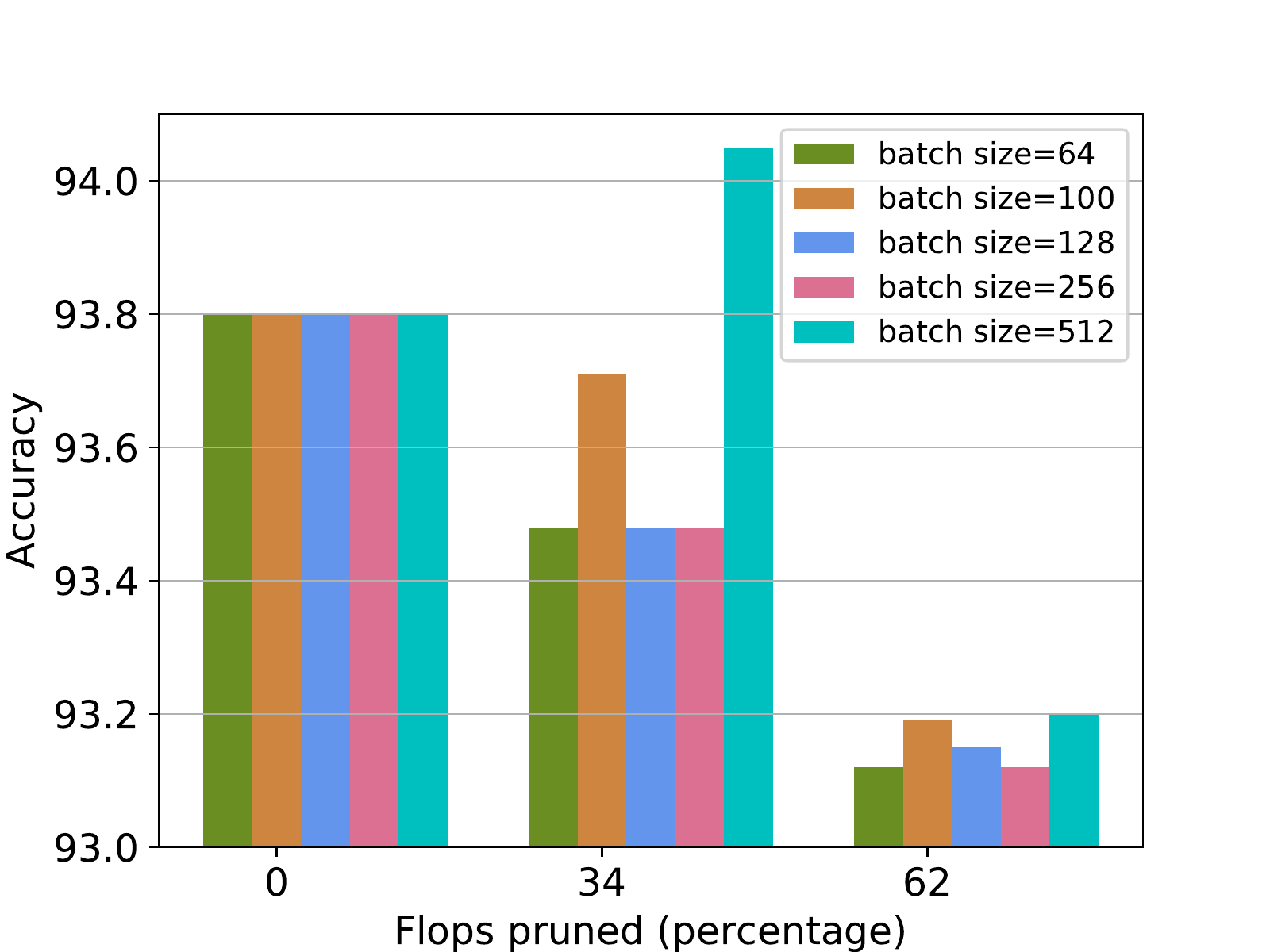} 
  \caption{Accuracy of ResNet-56 architecure on CIFAR10 dataset for various batchsizes.}
  \label{fig:batch_size}
  \end{figure}

\textcolor{myorange}{
\textit{2) Effect of batch size}:
Here, the effect of batch size during the computation of Relevance of filters is analyzed. Batch sizes of 64, 100, 128, 256, and 512 are used with ResNet-56 on CIFAR-10 dataset. The results obtained using the HRel method with a batch size of 100 for ResNet-56 are reported in Table \ref{tab:res56_results}. As illustrated in Fig. \ref{fig:batch_size}, it is observed that the batch size 64 resulted in comparatively lower accuracy. At the initial pruning iterations, the first and second-best performances are obtained with the batch size of 512 and 100, respectively. However, for higher pruned FLOPs, all the batch sizes produced similar results. Hence, the batch size during the estimation of filters' Relevance has less effect on the final performance of the HRel method.
}

\section{Conclusion}
\label{sec:conclusion}
In this paper, filters in CNNs are pruned based on their Relevance value. The Relevance measure is chosen based on IB theory, which is measured using the mutual information (MI) between the activations maps of the respective filters and the ground truths. The proposed HRel pruning method is evaluated on MNIST, CIFAR-10, and ImageNet datasets using LeNet-5, VGG-16, ResNet-56, ResNet-110, and ResNet-50 models. 
The pruning results obtained using the HRel method are superior compared to the current state-of-the-art pruning methods. The IP dynamics show the significance of the pruning criteria. The analysis of IP plane dynamics before and after pruning for the different CNNs suggests that the information loss after pruning is negligible. The filters’ Relevance is observed to increase from initial pruning iteration to final iteration except for ResNet-50 on ImageNet.
The deployment of lightweight models that are pruned using the HRel method on edge devices such as drones, mobiles may be potential future research.

\section*{Acknowledgements}
We thank Nvidia for donating two Titan X GPUs, which are used to perform the experiments of this research.

{\small
\bibliographystyle{ieee_fullname}
\bibliography{refs}

\begin{thebibliography}{10}\itemsep=-1pt

\bibitem{amjad2021understanding}
RA Amjad, K Liu, and BC Geiger.
\newblock Understanding neural networks and individual neuron importance via
  information-ordered cumulative ablation.
\newblock {\em IEEE Transactions on Neural Networks and Learning Systems},
  2021.

\bibitem{amodei2016deep}
Dario Amodei, Sundaram Ananthanarayanan, Rishita Anubhai, Jingliang Bai, Eric
  Battenberg, Carl Case, Jared Casper, Bryan Catanzaro, Qiang Cheng, Guoliang
  Chen, et~al.
\newblock Deep speech 2: End-to-end speech recognition in english and mandarin.
\newblock In {\em International conference on machine learning}, pages
  173--182. PMLR, 2016.

\bibitem{ayinde2019redundant}
Babajide~O Ayinde, Tamer Inanc, and Jacek~M Zurada.
\newblock Redundant feature pruning for accelerated inference in deep neural
  networks.
\newblock {\em Neural Networks}, 118:148--158, 2019.

\bibitem{balda2018information}
Emilio~Rafael Balda, Arash Behboodi, and Rudolf Mathar.
\newblock An information theoretic view on learning of artificial neural
  networks.
\newblock In {\em 2018 12th International Conference on Signal Processing and
  Communication Systems (ICSPCS)}, pages 1--8. IEEE, 2018.

\bibitem{basha2021deep}
SH Basha, Mohammad Farazuddin, Viswanath Pulabaigari, Shiv~Ram Dubey, and
  Snehasis Mukherjee.
\newblock Deep model compression based on the training history.
\newblock {\em arXiv preprint arXiv:2102.00160}, 2021.

\bibitem{belghazi2018mutual}
Mohamed~Ishmael Belghazi, Aristide Baratin, Sai Rajeshwar, Sherjil Ozair,
  Yoshua Bengio, Aaron Courville, and Devon Hjelm.
\newblock Mutual information neural estimation.
\newblock In {\em International Conference on Machine Learning}, pages
  531--540. PMLR, 2018.

\bibitem{botev2017nesterov}
Aleksandar Botev, Guy Lever, and David Barber.
\newblock Nesterov's accelerated gradient and momentum as approximations to
  regularised update descent.
\newblock In {\em 2017 International Joint Conference on Neural Networks
  (IJCNN)}, pages 1899--1903. IEEE, 2017.

\bibitem{chelombiev2019adaptive}
Ivan Chelombiev, Conor Houghton, and Cian O'Donnell.
\newblock Adaptive estimators show information compression in deep neural
  networks.
\newblock {\em arXiv preprint arXiv:1902.09037}, 2019.

\bibitem{cheng2015deep}
Zezhou Cheng, Qingxiong Yang, and Bin Sheng.
\newblock Deep colorization.
\newblock In {\em Proceedings of the IEEE International Conference on Computer
  Vision}, pages 415--423, 2015.

\bibitem{Courbariaux2015BinaryConnectTD}
Matthieu Courbariaux, Yoshua Bengio, and J. David.
\newblock Binaryconnect: Training deep neural networks with binary weights
  during propagations.
\newblock In {\em NIPS}, 2015.

\bibitem{10.5555/2969442.2969588}
Matthieu Courbariaux, Yoshua Bengio, and Jean-Pierre David.
\newblock Binaryconnect: Training deep neural networks with binary weights
  during propagations.
\newblock In {\em Proceedings of the 28th International Conference on Neural
  Information Processing Systems - Volume 2}, NIPS'15, page 3123–3131,
  Cambridge, MA, USA, 2015. MIT Press.

\bibitem{10.5555/1146355}
Thomas~M. Cover and Joy~A. Thomas.
\newblock {\em Elements of Information Theory (Wiley Series in
  Telecommunications and Signal Processing)}.
\newblock Wiley-Interscience, USA, 2006.

\bibitem{dai2018compressing}
Bin Dai, Chen Zhu, Baining Guo, and David Wipf.
\newblock Compressing neural networks using the variational information
  bottleneck.
\newblock In {\em International Conference on Machine Learning}, pages
  1135--1144. PMLR, 2018.

\bibitem{dong2019network}
Xuanyi Dong and Yi Yang.
\newblock Network pruning via transformable architecture search.
\newblock {\em arXiv preprint arXiv:1905.09717}, 2019.

\bibitem{fayek2017evaluating}
Haytham~M Fayek, Margaret Lech, and Lawrence Cavedon.
\newblock Evaluating deep learning architectures for speech emotion
  recognition.
\newblock {\em Neural Networks}, 92:60--68, 2017.

\bibitem{ganesh2020mint}
Madan~Ravi Ganesh, Jason~J Corso, and Salimeh~Yasaei Sekeh.
\newblock Mint: Deep network compression via mutual information-based neuron
  trimming.
\newblock {\em arXiv preprint arXiv:2003.08472}, 2020.

\bibitem{giraldo2014measures}
Luis Gonzalo~Sanchez Giraldo, Murali Rao, and Jose~C Principe.
\newblock Measures of entropy from data using infinitely divisible kernels.
\newblock {\em IEEE Transactions on Information Theory}, 61(1):535--548, 2014.

\bibitem{goldfeld2018estimating}
Ziv Goldfeld, Ewout van~den Berg, Kristjan Greenewald, Igor Melnyk, Nam Nguyen,
  Brian Kingsbury, and Yury Polyanskiy.
\newblock Estimating information flow in deep neural networks.
\newblock {\em arXiv preprint arXiv:1810.05728}, 2018.

\bibitem{guo2020dmcp}
Shaopeng Guo, Yujie Wang, Quanquan Li, and Junjie Yan.
\newblock Dmcp: Differentiable markov channel pruning for neural networks.
\newblock In {\em Proceedings of the IEEE/CVF Conference on Computer Vision and
  Pattern Recognition}, pages 1539--1547, 2020.

\bibitem{guyon2004result}
Isabelle Guyon, Steve Gunn, Asa Ben-Hur, and Gideon Dror.
\newblock Result analysis of the nips 2003 feature selection challenge.
\newblock {\em Advances in neural information processing systems}, 17, 2004.

\bibitem{han2015learning}
Song Han, Jeff Pool, John Tran, and William~J Dally.
\newblock Learning both weights and connections for efficient neural network.
\newblock In {\em NIPS}, 2015.

\bibitem{he2016deep}
Kaiming He, Xiangyu Zhang, Shaoqing Ren, and Jian Sun.
\newblock Deep residual learning for image recognition.
\newblock In {\em Proceedings of the IEEE conference on computer vision and
  pattern recognition}, pages 770--778, 2016.

\bibitem{he2020learning}
Yang He, Yuhang Ding, Ping Liu, Linchao Zhu, Hanwang Zhang, and Yi Yang.
\newblock Learning filter pruning criteria for deep convolutional neural
  networks acceleration.
\newblock In {\em Proceedings of the IEEE/CVF conference on computer vision and
  pattern recognition}, pages 2009--2018, 2020.

\bibitem{he2019asymptotic}
Yang He, Xuanyi Dong, Guoliang Kang, Yanwei Fu, Chenggang Yan, and Yi Yang.
\newblock Asymptotic soft filter pruning for deep convolutional neural
  networks.
\newblock {\em IEEE transactions on cybernetics}, 50(8):3594--3604, 2019.

\bibitem{he2018soft}
Y He, G Kang, X Dong, Y Fu, and Y Yang.
\newblock Soft filter pruning for accelerating deep convolutional neural
  networks.
\newblock In {\em IJCAI International Joint Conference on Artificial
  Intelligence}, 2018.

\bibitem{he2019filter}
Yang He, Ping Liu, Ziwei Wang, Zhilan Hu, and Yi Yang.
\newblock Filter pruning via geometric median for deep convolutional neural
  networks acceleration.
\newblock In {\em Proceedings of the IEEE/CVF Conference on Computer Vision and
  Pattern Recognition}, pages 4340--4349, 2019.

\bibitem{44873}
Geoffrey Hinton, Oriol Vinyals, and Jeffrey Dean.
\newblock Distilling the knowledge in a neural network.
\newblock In {\em NIPS Deep Learning and Representation Learning Workshop},
  2015.

\bibitem{hu2016network}
Hengyuan Hu, Rui Peng, Yu-Wing Tai, and Chi-Keung Tang.
\newblock Network trimming: A data-driven neuron pruning approach towards
  efficient deep architectures.
\newblock {\em arXiv preprint arXiv:1607.03250}, 2016.

\bibitem{journals/corr/JaderbergVZ14}
Max Jaderberg, Andrea Vedaldi, and Andrew Zisserman.
\newblock Speeding up convolutional neural networks with low rank expansions.
\newblock {\em CoRR}, abs/1405.3866, 2014.

\bibitem{jonsson2020convergence}
Hlynur J{\'o}nsson, Giovanni Cherubini, and Evangelos Eleftheriou.
\newblock Convergence behavior of dnns with mutual-information-based
  regularization.
\newblock {\em Entropy}, 22(7):727, 2020.

\bibitem{jordao2020deep}
Artur Jordao, Fernando Yamada, and William~Robson Schwartz.
\newblock Deep network compression based on partial least squares.
\newblock {\em Neurocomputing}, 406:234--243, 2020.

\bibitem{kolchinsky2017estimating}
Artemy Kolchinsky and Brendan~D Tracey.
\newblock Estimating mixture entropy with pairwise distances.
\newblock {\em Entropy}, 19(7):361, 2017.

\bibitem{kraskov2004estimating}
Alexander Kraskov, Harald St{\"o}gbauer, and Peter Grassberger.
\newblock Estimating mutual information.
\newblock {\em Physical review E}, 69(6):066138, 2004.

\bibitem{krizhevsky2009learning}
Alex Krizhevsky, Geoffrey Hinton, et~al.
\newblock Learning multiple layers of features from tiny images.
\newblock 2009.

\bibitem{lecun-mnisthandwrittendigit-2010}
Yann LeCun and Corinna Cortes.
\newblock {MNIST} handwritten digit database.
\newblock 2010.

\bibitem{lecun1990optimal}
Yann LeCun, John~S Denker, and Sara~A Solla.
\newblock Optimal brain damage.
\newblock In {\em Advances in neural information processing systems}, pages
  598--605, 1990.

\bibitem{lee2020channel}
Min~Kyu Lee, Seunghyun Lee, Sang~Hyuk Lee, and Byung~Cheol Song.
\newblock Channel pruning via gradient of mutual information for light-weight
  convolutional neural networks.
\newblock In {\em 2020 IEEE International Conference on Image Processing
  (ICIP)}, pages 1751--1755. IEEE, 2020.

\bibitem{leonenko2008class}
Nikolai Leonenko, Luc Pronzato, Vippal Savani, et~al.
\newblock A class of r{\'e}nyi information estimators for multidimensional
  densities.
\newblock {\em Annals of statistics}, 36(5):2153--2182, 2008.

\bibitem{li2016pruning}
Hao Li, Asim Kadav, Igor Durdanovic, Hanan Samet, and Hans~Peter Graf.
\newblock Pruning filters for efficient convnets.
\newblock {\em arXiv preprint arXiv:1608.08710}, 2016.

\bibitem{lin2020hrank}
Mingbao Lin, Rongrong Ji, Yan Wang, Yichen Zhang, Baochang Zhang, Yonghong
  Tian, and Ling Shao.
\newblock Hrank: Filter pruning using high-rank feature map.
\newblock In {\em Proceedings of the IEEE/CVF Conference on Computer Vision and
  Pattern Recognition}, pages 1529--1538, 2020.

\bibitem{lin2020channel}
Mingbao Lin, Rongrong Ji, Yuxin Zhang, Baochang Zhang, Yongjian Wu, and
  Yonghong Tian.
\newblock Channel pruning via automatic structure search.
\newblock In {\em Proceedings of the International Joint Conference on
  Artificial Intelligence (IJCAI)}, pages 673 -- 679, 2020.

\bibitem{lin2019towards}
Shaohui Lin, Rongrong Ji, Chenqian Yan, Baochang Zhang, Liujuan Cao, Qixiang
  Ye, Feiyue Huang, and David Doermann.
\newblock Towards optimal structured cnn pruning via generative adversarial
  learning.
\newblock In {\em Proceedings of the IEEE/CVF Conference on Computer Vision and
  Pattern Recognition}, pages 2790--2799, 2019.

\bibitem{liu2017learning}
Zhuang Liu, Jianguo Li, Zhiqiang Shen, Gao Huang, Shoumeng Yan, and Changshui
  Zhang.
\newblock Learning efficient convolutional networks through network slimming.
\newblock In {\em Proceedings of the IEEE international conference on computer
  vision}, pages 2736--2744, 2017.

\bibitem{liu2019metapruning}
Zechun Liu, Haoyuan Mu, Xiangyu Zhang, Zichao Guo, Xin Yang, Kwang-Ting Cheng,
  and Jian Sun.
\newblock Metapruning: Meta learning for automatic neural network channel
  pruning.
\newblock In {\em Proceedings of the IEEE/CVF International Conference on
  Computer Vision}, pages 3296--3305, 2019.

\bibitem{luo2017thinet}
Jian-Hao Luo, Jianxin Wu, and Weiyao Lin.
\newblock Thinet: A filter level pruning method for deep neural network
  compression.
\newblock In {\em Proceedings of the IEEE international conference on computer
  vision}, pages 5058--5066, 2017.

\bibitem{min20182pfpce}
Chuhan Min, Aosen Wang, Yiran Chen, Wenyao Xu, and Xin Chen.
\newblock 2pfpce: Two-phase filter pruning based on conditional entropy.
\newblock {\em arXiv preprint arXiv:1809.02220}, 2018.

\bibitem{muresan2018fruit}
Horea MURESAN and Mihai OLTEAN.
\newblock Fruit recognition from images using deep learning.
\newblock {\em Acta Univ. Sapientiae}, 10(1):26--42, 2018.

\bibitem{nielsen2002quantum}
Michael~A Nielsen and Isaac Chuang.
\newblock Quantum computation and quantum information, 2002.

\bibitem{noshad2019scalable}
Morteza Noshad, Yu Zeng, and Alfred~O Hero.
\newblock Scalable mutual information estimation using dependence graphs.
\newblock In {\em ICASSP 2019-2019 IEEE International Conference on Acoustics,
  Speech and Signal Processing (ICASSP)}, pages 2962--2966. IEEE, 2019.

\bibitem{purwani2017analyzing}
Sri Purwani, Julita Nahar, and Carole Twining.
\newblock Analyzing bin-width effect on the computed entropy.
\newblock In {\em AIP Conference Proceedings}, volume 1868, page 040008. AIP
  Publishing LLC, 2017.

\bibitem{redmon2017yolo9000}
Joseph Redmon and Ali Farhadi.
\newblock Yolo9000: better, faster, stronger.
\newblock In {\em Proceedings of the IEEE conference on computer vision and
  pattern recognition}, pages 7263--7271, 2017.

\bibitem{ren2015faster}
Shaoqing Ren, Kaiming He, Ross Girshick, and Jian Sun.
\newblock Faster r-cnn: Towards real-time object detection with region proposal
  networks.
\newblock {\em arXiv preprint arXiv:1506.01497}, 2015.

\bibitem{Romero2015FitNetsHF}
A. Romero, Nicolas Ballas, S. Kahou, Antoine Chassang, C. Gatta, and Yoshua
  Bengio.
\newblock Fitnets: Hints for thin deep nets.
\newblock {\em CoRR}, abs/1412.6550, 2015.

\bibitem{russakovsky2015imagenet}
Olga Russakovsky, Jia Deng, Hao Su, Jonathan Krause, Sanjeev Satheesh, Sean Ma,
  Zhiheng Huang, Andrej Karpathy, Aditya Khosla, Michael Bernstein, et~al.
\newblock Imagenet large scale visual recognition challenge.
\newblock {\em International journal of computer vision}, 115(3):211--252,
  2015.

\bibitem{saxe2019information}
Andrew~M Saxe, Yamini Bansal, Joel Dapello, Madhu Advani, Artemy Kolchinsky,
  Brendan~D Tracey, and David~D Cox.
\newblock On the information bottleneck theory of deep learning.
\newblock {\em Journal of Statistical Mechanics: Theory and Experiment},
  2019(12):124020, 2019.

\bibitem{shwartz2017opening}
Ravid Shwartz-Ziv and Naftali Tishby.
\newblock Opening the black box of deep neural networks via information.
\newblock {\em arXiv preprint arXiv:1703.00810}, 2017.

\bibitem{silverman1986monographs}
Bernard~W Silverman.
\newblock Monographs on statistics and applied probability.
\newblock {\em Density estimation for statistics and data analysis}, 26, 1986.

\bibitem{simonyan2014very}
Karen Simonyan and Andrew Zisserman.
\newblock Very deep convolutional networks for large-scale image recognition.
\newblock {\em arXiv preprint arXiv:1409.1556}, 2014.

\bibitem{singh2020leveraging}
Pravendra Singh, Vinay~Kumar Verma, Piyush Rai, and Vinay Namboodiri.
\newblock Leveraging filter correlations for deep model compression.
\newblock In {\em Proceedings of the IEEE/CVF Winter Conference on Applications
  of Computer Vision}, pages 835--844, 2020.

\bibitem{singh2020acceleration}
Pravendra Singh, Vinay~Kumar Verma, Piyush Rai, and Vinay~P Namboodiri.
\newblock Acceleration of deep convolutional neural networks using adaptive
  filter pruning.
\newblock {\em IEEE Journal of Selected Topics in Signal Processing},
  14(4):838--847, 2020.

\bibitem{su2020locally}
Xiu Su, Shan You, Tao Huang, Fei Wang, Chen Qian, Changshui Zhang, and Chang
  Xu.
\newblock Locally free weight sharing for network width search.
\newblock In {\em International Conference on Learning Representations}, 2020.

\bibitem{tapia2020information}
Nicol{\'a}s~I Tapia and Pablo~A Est{\'e}vez.
\newblock On the information plane of autoencoders.
\newblock In {\em 2020 International Joint Conference on Neural Networks
  (IJCNN)}, pages 1--8. IEEE, 2020.

\bibitem{tishby2000information}
Naftali Tishby, Fernando~C Pereira, and William Bialek.
\newblock The information bottleneck method.
\newblock {\em arXiv preprint physics/0004057}, 2000.

\bibitem{wang2017accurate}
Sheng Wang, Siqi Sun, Zhen Li, Renyu Zhang, and Jinbo Xu.
\newblock Accurate de novo prediction of protein contact map by ultra-deep
  learning model.
\newblock {\em PLoS computational biology}, 13(1):e1005324, 2017.

\bibitem{wang2019cop}
Wenxiao Wang, Cong Fu, Jishun Guo, Deng Cai, and Xiaofei He.
\newblock Cop: Customized deep model compression via regularized
  correlation-based filter-level pruning.
\newblock {\em arXiv preprint arXiv:1906.10337}, 2019.

\bibitem{wen2020structured}
Liangjian Wen, Xuanyang Zhang, Haoli Bai, and Zenglin Xu.
\newblock Structured pruning of recurrent neural networks through neuron
  selection.
\newblock {\em Neural Networks}, 123:134--141, 2020.

\bibitem{wickstrom2019information}
Kristoffer Wickstr{\o}m, Sigurd L{\o}kse, Michael Kampffmeyer, Shujian Yu, Jose
  Principe, and Robert Jenssen.
\newblock Information plane analysis of deep neural networks via matrix-based
  renyi's entropy and tensor kernels.
\newblock {\em arXiv preprint arXiv:1909.11396}, 2019.

\bibitem{Williams2010NonnegativeDO}
P.~L. Williams and R. Beer.
\newblock Nonnegative decomposition of multivariate information.
\newblock {\em ArXiv}, abs/1004.2515, 2010.

\bibitem{Wu2016QuantizedCN}
J. Wu, C. Leng, Yuhang Wang, Q. Hu, and J. Cheng.
\newblock Quantized convolutional neural networks for mobile devices.
\newblock {\em 2016 IEEE Conference on Computer Vision and Pattern Recognition
  (CVPR)}, pages 4820--4828, 2016.

\bibitem{yasaei2019geometric}
Salimeh Yasaei~Sekeh and Alfred~O Hero.
\newblock Geometric estimation of multivariate dependency.
\newblock {\em Entropy}, 21(8):787, 2019.

\bibitem{yu2019autoslim}
Jiahui Yu and Thomas Huang.
\newblock Autoslim: Towards one-shot architecture search for channel numbers.
\newblock {\em arXiv preprint arXiv:1903.11728}, 2019.

\bibitem{yu2019multivariate}
Shujian Yu, Luis Gonzalo~Sanchez Giraldo, Robert Jenssen, and Jose~C. Principe.
\newblock Multivariate extension of matrix-based renyi's $\alpha$-order entropy
  functional, 2019.

\bibitem{yu2020understanding}
Shujian Yu, Kristoffer Wickstr{\o}m, Robert Jenssen, and Jos{\'e}~C
  Pr{\'\i}ncipe.
\newblock Understanding convolutional neural networks with information theory:
  An initial exploration.
\newblock {\em IEEE transactions on neural networks and learning systems},
  2020.

\end{thebibliography}
}

\end{document}